**RESEARCH ARTICLE**

# Toward Physically Plausible Data-Driven Models: A Novel Neural Network Approach to Symbolic Regression

JIŘÍ KUBALÍK[1], ERIK DERNER[1], AND ROBERT BABUŠKA[1,2], (Member, IEEE)
[1]Czech Institute of Informatics, Robotics, and Cybernetics, Czech Technical University in Prague, 16000 Prague, Czech Republic
[2]Department of Cognitive Robotics, Delft University of Technology, 2628 CD Delft, The Netherlands
Corresponding author: Jiří Kubalík (jiri.kubalik@cvut.cz)

This work was supported by the European Regional Development Fund under the Project Robotics for Industry 4.0 under Grant CZ.02.1.01/0.0/0.0/15_003/0000470.

**ABSTRACT** Many real-world systems can be described by mathematical models that are human-comprehensible, easy to analyze and help explain the system's behavior. Symbolic regression is a method that can automatically generate such models from data. Historically, symbolic regression has been predominantly realized by genetic programming, a method that evolves populations of candidate solutions that are subsequently modified by genetic operators crossover and mutation. However, this approach suffers from several deficiencies: it does not scale well with the number of variables and samples in the training data – models tend to grow in size and complexity without an adequate accuracy gain, and it is hard to fine-tune the model coefficients using just genetic operators. Recently, neural networks have been applied to learn the whole analytic model, i.e., its structure and the coefficients, using gradient-based optimization algorithms. This paper proposes a novel neural network-based symbolic regression method that constructs physically plausible models based on even very small training data sets and prior knowledge about the system. The method employs an adaptive weighting scheme to effectively deal with multiple loss function terms and an epoch-wise learning process to reduce the chance of getting stuck in poor local optima. Furthermore, we propose a parameter-free method for choosing the model with the best interpolation and extrapolation performance out of all the models generated throughout the whole learning process. We experimentally evaluate the approach on four test systems: the TurtleBot 2 mobile robot, the magnetic manipulation system, the equivalent resistance of two resistors in parallel, and the longitudinal force of the anti-lock braking system. The results clearly show the potential of the method to find parsimonious models that comply with the prior knowledge provided.

## I. INTRODUCTION

Symbolic regression (SR) is a data-driven method that generates models in the form of analytic formulas. It has been successfully used in many nonlinear modeling tasks with quite impressive results [1], [2], [3], [4]. Historically, SR has been predominantly realized using genetic programming (GP) [1], [5], [6], [7], [8], a method that evolves a population of candidate solutions for a number of generations.

The associate editor coordinating the review of this manuscript and approving it for publication was Chao Tong.

This gradient-free learning process is driven by a selection strategy that prefers high-quality solutions to poor ones. New candidate solutions are created by applying crossover and mutation genetic operators. Some GP-based approaches use the loss function gradient to fine-tune the inner coefficients of the model [9], [10], [11].

SR has several advantages over other data-driven modeling methods. For example, contrary to (deep) neural networks, which belong to data-hungry approaches, SR can construct good models even from very small training data sets [12], [13]. SR is also suitable for incorporating prior knowledge about the desired properties of the modeled system [8], [14], [15]. This is very important, especially when the data set does not sufficiently cover the input space or when some parts of the input space are entirely missing from the data set. Even when a sufficiently large and informative training set is available, methods that minimize only the training error tend to yield partially incorrect models, for instance, in terms of their steady-state characteristics and local or even global behavior.

Despite their high popularity, GP-based SR methods suffer from several deficiencies. The models tend to increase in size and complexity without an adequate increase in performance. This phenomenon is known as code bloat [16]. Furthermore, GP-based approaches do not scale well with the number of variables and samples in the training data set. This is because an entire population of formulas has to be evolved and evaluated repeatedly through many generations. Last but not least, it is hard to tune the coefficients of the models using just genetic operators.

To cope with the issues mentioned above, several approaches have recently been proposed for using neural networks (NN) to learn analytic formulas using gradient-based optimization algorithms [17], [18], [19], [20], [21], [22]. All these approaches share the idea that analytic models can be represented by a heterogeneous NN with units implementing mathematical operators and elementary functions, such as {+, −, ∗, /, sin, exp, etc.}. The network weights are adjusted using standard gradient-based methods with the ultimate goal of minimizing the training error while maximally reducing the number of active units.[1] The final NN then represents a parsimonious analytic formula. The individual approaches differ in how the learning process is driven toward obtaining the final model.

In this paper, we propose a novel NN-based SR approach, N4SR (pronounced as "enfɔ:sə"), that allows for using prior knowledge represented by and evaluated on constraint samples as introduced in [14] and [15]. The NN uses an EQL-like architecture [17] with skip connections [20]. The learning process is divided into epochs. During the learning process, models of varying sizes, measured by the number of active weights, are generated. Then, the final model is chosen as the best-performing model among the least complex ones. The model's performance is judged based on its validation root-mean-square error and compliance with the validity constraints representing the prior knowledge. To the best of our knowledge, this is the first paper proposing incorporating prior knowledge into the NN for the SR task of generating compact analytical formulas from data.

The problem of seeking a sparse model that has a low training error and exhibits desired characteristics is a multi-objective optimization problem. This implies a strong interplay among the respective terms of the loss function. Some terms may become dominant in the loss function while suppressing the effect of some other terms. To remedy this problem, we introduce a self-adaptive strategy to keep the pair-wise ratios between the terms around the required values during optimization. In summary, the main contributions of this paper are:

- We introduce a neural network-based approach to symbolic regression that uses training data and prior knowledge to generate precise and, at the same time, physically plausible models.
- We propose a self-adaptive strategy to control the contributions of the training error term, the regularization term, and the constraint error term during the optimization process. We show that this method is effective compared to the non-adaptive one. Moreover, it reduces the number of parameters to be tuned for each SR instance.
- We propose the final model selection method based on the model's complexity and the constraint violation error. The method does not require any extrapolation test set. We show that the proposed method is competitive with the variant based on the extrapolation error, which needs data sampled from the extrapolation domain.
- The proposed N4SR was thoroughly evaluated on four test problems, including the validation of our design choices, and compared to other relevant methods.

The paper is organized as follows. Section II surveys the related work. Then, the particular SR problem considered in this work is defined in Section III. The proposed method is described in Section IV. Section V presents the experiments and discusses the results obtained. Finally, Section VI concludes the paper.

## II. RELATED WORK

One of the first works on using NN for symbolic regression is [17], where the Equation Learner (EQL) was introduced. It works with a simple feed-forward multi-layer architecture with several unit types – sigmoid, sine, cosine, identity, and multiplication. The network is trained using a stochastic gradient descent algorithm with mini-batches, Adam [23], and a Lasso-like objective combining the $L_2$ training loss and $L_1$ regularization. Moreover, it uses a hybrid regularization strategy starting with several update steps without regularization, followed by a regularization phase to enforce a sparse network structure to emerge. In the final phase, regularization is disabled, but the $L_0$ norm of the weights is still enforced, i.e., all weights close to 0 are set to 0. An important question is choosing the right network (i.e., the final model) among all the network instances generated during the learning process. In [17], they solve it by ranking the network instances w.r.t. validation error and sparsity and selecting the one with the smallest $L_2$ norm (in rank-space). However, it was shown that in some cases, this does not select a network instance with the best performance metrics.

In [18], an extended version of EQL, denoted as EQL$^{\div}$, was proposed. In addition to the original EQL, it allows for modeling divisions using a modified architecture that places the division units in the output layer. The objective function is a linear combination of $L_2$ training loss and $L_1$ regularization extended by a penalty term for invalid denominator values $P^\theta$. Furthermore, special *penalty epochs* are injected at regular intervals into the training process to prevent output values on data from extrapolation regions having a very different magnitude than the outputs observed on the training data. During the penalty epochs, only the penalty function $P^\theta + P^{bound}$ is minimized, where $P^{bound}$ penalizes outputs larger than the maximal desired value observed on all data points (including the extrapolation ones). In this way, a reasonable but not necessarily correct behavior of the model in the extrapolation region is enforced. Moreover, one must estimate the maximal desired output value in advance, which cannot be done reliably in general. Here, the model selection method chooses the network instance that minimizes the sum of normalized interpolation and extrapolation validation errors, where the extrapolation error is calculated on several measured extrapolation points. On the one hand, this method was shown to work better than the one used in EQL, on the other hand, it still relies on known extrapolation points, though just a few.

---
[1]A unit is considered active if and only if it has at least one above-threshold input weight and it contributes to the NN output.

In [21], an EQL architecture with other deep learning architectures and $L_{0.5}$ regularization was proposed. Its power was demonstrated on a simple arithmetic task where the EQL learns how to ''add'' two numbers that are extracted from handwritten MNIST[2] digit images by a convolutional network, and on a set of experiments, where the EQL network was applied to analyze physical time-varying systems. Partially inspired by the EQL network, a new multi-layer NN architecture, OccamNet, that represents a probability distribution over functions was proposed in [20]. It uses skip-connections similar to those in DenseNet [24] and a temperature-controlled connectivity scheme. It uses the probabilistic interpretation of the softmax function by sampling sparse paths through a network to maximize sparsity. The Mathematical Operation Network (MathONet) proposed in [22] also uses EQL-like NN architecture. The sparse sub-graph of the NN is sought using a Bayesian learning approach that incorporates structural and non-structural sparsity priors. The system was shown to be able to discover ordinary and partial differential equations from the observations.

A different class of NN-based SR approaches uses deep neural network transformers such as the GPT-2 [25]. They learn the transformer model using a large amount of training data where each sample is typically a tuple of the form (formula, data sampled from the formula). During inference, the transformer model constructs the formula based on a particular data set query. This approach has its advantages and disadvantages, but this is out of the scope of this work. We refer interested readers to [26], [27], [28], and [29].

Note that no prior knowledge has been used in the approaches above. When learning the model, these methods focus purely on minimizing the error computed on the training data. Consequently, they are likely to produce physically inconsistent results and exhibit a rather low ability to generalize to out-of-training-distribution samples. This issue is addressed by the recent trend observed in the literature where an increasing research effort has been focused on integrating traditional physics-based modeling approaches with machine learning techniques [30], [31]. In the context of this work, an interesting approach is the physics-informed neural network framework that trains deep neural networks to solve supervised learning tasks with relatively small amounts of training data and the underlying physics described by partial differential equations [32], [33].

Combining SR with prior knowledge is still a rather new research topic. In [8], the Counterexample-Driven Symbolic Regression based on the Counterexample-Driven Genetic Programming [34] was introduced. It represents domain knowledge as a set of constraints expressed as logical formulas. A Satisfiability Modulo Theories (SMT) solver is used to verify whether a given model meets the constraints. A similar approach called Logic Guided Genetic Algorithms (LGGA) was proposed in [35]. Here, the domain-specific knowledge is called auxiliary truths (AT), which are simple mathematical facts known a priori about the unknown function sought. ATs are represented as mathematical formulas. Candidate models are evaluated using a weighted sum of the classical training mean squared error and the truth error, which measures how much the model violates given ATs. Unlike CDSR, LGGA performs computationally light consistency checks via ATs' formulas evaluations on the data set.

In our previous work [14], [15], we proposed a multi-objective approach that optimizes models with respect to the training accuracy and the level of compliance with prior knowledge simultaneously. It assumes that prior knowledge can be written as nonlinear inequality or equality constraint that the system must obey. Synthetic constraint samples (i.e., samples not measured on the system) are generated specifically for each constraint, and the desired inequality or equality relation is defined on them. Then, the constraint violation error measures how much the model violates the desired inequality or equality relations over the constraint samples.

The informed EQL (iEQL) proposed in [19] is another extension of EQL. It uses expert knowledge about permitted or prohibited equation components and a domain-dependent structured sparsity prior. In artificial and real-world experiments, the authors demonstrated that iEQL could learn interpretable models of high predictive power. However, this type of expert knowledge might be hard to define reliably for some problems, as even nontrivial nested structures may be beneficial in some cases. Shape-constrained symbolic regression was introduced in [36] and [37], which allows to include vague prior knowledge by measuring properties of the shape of the model, such as monotonicity and convexity, using interval arithmetic. It has been shown that introducing shape constraints helps find more realistic models.

---

[2]http://yann.lecun.com/exdb/mnist/

## III. PROBLEM DEFINITION

In this work, we consider a regression problem where a neural network model representing the function $\phi : \mathbb{R}^n \to \mathbb{R}$ is sought, with $n$ being the number of input variables. In particular, a maximally sparse neural network model representing a desirably concise analytic expression is sought that

- minimizes the error observed on the training data set,
- maximizes its validity (see below),
- maximizes its compliance with the constraints imposed on the model, which define the desired properties of the model,
- and exhibits good extrapolation performance at the same time.

The model's validity reflects that the neural network may contain units with singularities, such as the division $a/b$, which exhibits a singularity at $b = 0$. In general, multiple types of singularity units can be considered here. We denote the set of singularity types used in the network as $T^s$. The constraint satisfaction measure is calculated on a set of samples generated specifically for each constraint type as proposed in [14] and [15]. The set of all constraints defined for the particular SR problem is denoted as $T^c$. The sparsity of the model is measured by the number of active weights and units in the neural network. The model's extrapolation performance is evaluated on a test data set that samples regions of the input space, either entirely omitted or present very scarcely in the training data.

## IV. METHOD

This section describes the main components of N4SR, namely the architecture of the neural network, the forms of the loss function used in different stages of the training run, the self-adaptive loss terms weighting scheme, the learning process procedure, and the final model selection rule.

The learning process proceeds in iterations, where the tunable weights are adjusted by gradient descent. Each such iteration produces a NN instance with its specific parameters.

### A. DATA SETS USED FOR LEARNING

Before we describe the method itself, we introduce the following data sets used to train the neural network model:

- Training data, $D_t$ – contains data samples of the form $\mathbf{d}_i = (\mathbf{x}_i, y_i)$, where $\mathbf{x}_i \in \mathbb{R}^n$ is sampled from the training domain $\mathbb{D}_t$.
- Validation data, $D_v$ – contains data samples of the same form as $D_t$ sampled from $\mathbb{D}_t$ such that $D_v \cap D_t = \emptyset$. This data set is used to choose intermediate models within the learning process, see Section IV-F, and to choose the final model according to the proposed model selection method, see Section IV-E.
- Constraint data, $D_c$ – we adopt the constraint representation and evaluation scheme as proposed in [14]. The set $D_c$ contains the synthetic constraint samples generated for each constraint in $T^c$ on which the constraint violation errors will be calculated.

### B. ARCHITECTURE

The approach we propose in this paper uses an EQL-like architecture similar to the ones introduced in [19] and [20]. It takes advantage of the skip connections so that simple structures present in shallow layers can be efficiently learned due to a direct propagation of the error gradients. Moreover, these shallow structures can be refined and reused in the subsequent layers. We also allow for using units with singularities. Contrary to the EQL$^{\div}$, singularity units can be used at any layer of the network.

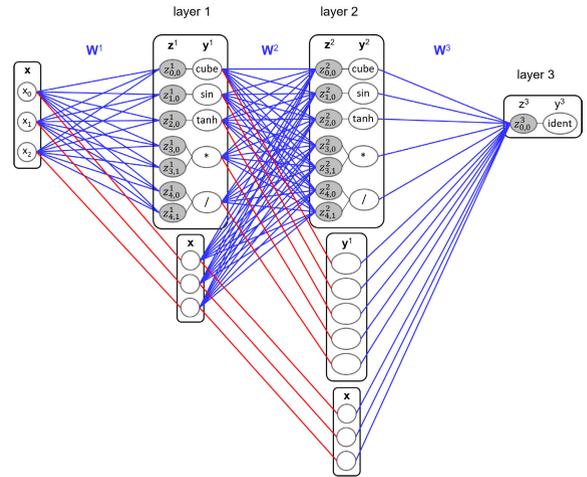

**FIGURE 1.** Network architecture with two hidden layers and one output layer unit. The blue lines mark links with learnable weights. The red lines are the skip connections leading from the source units of layer $k - 1$ to the copy units in layer $k$. These links are permanently set to 1. For simplicity, this scheme does not show the bias links leading to every $z$ node.

Figure 1 shows the core architecture components on an example of the neural network with two hidden layers and one output layer unit. Each hidden layer contains *learnable units* and the *copy units* (the term introduced in [19]). The learnable units are units whose input weights can be tuned within the learning process (i.e., the links shown in blue). The *learnable weights* of all learnable units are collected in the set $W_l$. The copy units in layer $k$ are copies of all units from the previous layer $k - 1$. Their weights are permanently set to 1 and are not subject to the learning process (these are the links shown in red). Units represent elementary unary functions, e.g., sin, cube, tanh, and binary functions (operators) such as multiplication $*$ and division $/$. Each learnable unit $i$ in layer $l$ calculates its output as

$$y_i^l = g(z_{i,0}^l), \text{ for unary elementary function } g \quad (1)$$

and

$$y_i^l = h(z_{i,0}^l, z_{i,1}^l), \text{ for binary elementary function } h \quad (2)$$

where $z_{i,\_}^l$ is an affine transformation of the whole previous layer's output $\mathbf{y}^{l-1}$ calculated as

$$z_{i,\_}^l = \mathbf{W}_i^l \mathbf{y}^{l-1} + b_i^l \quad (3)$$

with learnable weights $\mathbf{W}_i^l$ and bias $b_i^l$.

In Figure 1, each hidden layer contains learnable units composed of a single instance of each unary and binary unit type. In general, there may be multiple instances of each unit type. Similarly, the output layer may have multiple output units, not just a single one, and they can be of any type out of the unary and binary unit types, not necessarily the identity one as used here. This depends on the SR problem and on the expert knowledge about the formula form sought.

### C. LOSS FUNCTIONS

Throughout the learning process, the following three loss functions are used in the different stages of the learning process, described in Section IV-F.

$$\begin{aligned} \mathcal{L}_1 &= \mathcal{L}^t + \mathcal{L}^s \\ \mathcal{L}_2 &= \mathcal{L}^t + \mathcal{L}^s + \mathcal{L}^c \\ \mathcal{L}_3 &= \mathcal{L}^t + \mathcal{L}^s + \mathcal{L}^c + \mathcal{L}^r \end{aligned} \quad (4)$$

The loss functions are composed of the following terms:

- Training RMSE, $\mathcal{L}^t$ – this is the root-mean-square error (RMSE) observed on the training data set $D_t$.
- Singularity units loss, $\mathcal{L}^s$ – this term is defined as the weighted sum of scaled RMSE values $\rho^s_{j,k}$ calculated for each singularity unit type over the aggregated data set $D = D_t \cup D_v \cup D_c$. Note that all these data samples can be used to check the validity of the model as we do not need to know the required output value $y$. Instead, just the values of the respective $z$ node of each singularity unit are checked whether they take on values greater than or equal to a user-defined threshold $\theta^s_j$ for the given singularity type $j$. The $\mathcal{L}^s$ loss in the $k^{\text{th}}$ iteration is formally defined as

$$\begin{aligned} \mathcal{L}^s &= \alpha \sum_{j \in T^s} \frac{\rho^s_{j,k}}{h^s_{j,k}}, \\ \rho^s_{j,k} &= \sqrt{\frac{1}{|S_j||D|} \sum_{u \in S_j} \sum_{d \in D} (\mathrm{m}_{j,\mathrm{d}}(\theta^s_j, z_{u,d}))^2}, \\ h^s_{j,k} &= \frac{1}{N_w} \sum_{i=0}^{N_w - 1} \rho^s_{j,k-i}, \end{aligned} \quad (5)$$

where $S_j$ is the set of all singularity units of the given type $j$ in the model, $\mathrm{m}_{j,\mathrm{d}}(\theta^s_j, z_u)$ is a function defining a suitable error metric for the given singularity type, see below, and $z_u$ is the critical $z$ node of the respective singularity unit $u$ (e.g., the denominator in case of the division unit). Each $\rho^s_{j,k}$ value is divided by $h^s_{j,k}$, which is a scaling coefficient calculated as the mean of $\rho^s_j$ values observed during the last $N_w$ iterations. This kind of normalization is used to make the raw singularity terms $\rho^s_{j,k}$ contribute to the overall $\mathcal{L}^s$ with values of the same magnitude and typically close to one. Thus, if $\rho^s_{j,k}$ is less than $h^s_{j,k}$, then the singularity unit type $j$ contributes to $\mathcal{L}^s$ with a value less than one and vice versa. Without this normalization, some singularity type may dominate within the $\mathcal{L}^s$ term if its values are by orders of magnitude higher than the others. The coefficient $\alpha$ is determined using a self-adaptive scheme described below.

We adopt the implementation of the division operation $a/b$ originally proposed in [18] and further extended in [19] that uses the following error metric

$$\mathrm{m}_{j,\mathrm{d}}(\theta^s_j, z_{u,d}) = \begin{cases} \max(\theta^s_j - z_{u,d}, 0) & \text{if } z_{u,d} \neq 0 \\ 10 & \text{if } z_{u,d} = 0 \end{cases} \quad (6)$$

It makes use of the fact that real systems do not operate at the pole $b \to 0$; thus, only the positive branch of the hyperbola, $1/b$ where $b > 0$, is sufficient to represent the division while the numerator $a \in \mathbb{R}$ determines the sign of the division output value. Here, $b = z_u$ and the threshold $\theta^s_j$ is used to prevent $z_u$ values from converging very close to the pole value. If $z_{u,d} = 0$ for the given data sample, then it returns some large constant value to distinguish between cases where $z_{u,d}$ is close to becoming invalid and the case where it already has the invalid value (we use the constant of 10 in this work). This representation can also be used for other units exhibiting a singularity such as log which is defined on the interval $(0, \infty)$ with the singularity at 0.

- Constraint loss, $\mathcal{L}^c$ – this term accumulates the error of the model in terms of the prior knowledge violation. Like $\mathcal{L}^s$, this term is calculated as the weighted sum of scaled RMSE values calculated for each constraint on its own specific constraint data set. Formally, $\mathcal{L}^c$ is defined as

$$\mathcal{L}^c = \beta \sum_{j \in T^c} \frac{\rho^c_{j,k}}{h^c_{j,k}} \quad (7)$$

where $\rho^c_{j,k}$ is the root-mean-square error calculated for the $j^{\text{th}}$ constraint on its constraint data set and $h^c_{j,k}$ is the mean of historical $\rho^c_j$ values. The logic of equation (7) is analogous to the one described for equation (5). Note that constraint data samples are not regular data samples. Depending on the constraint type, a constraint sample is a tuple containing one or more unlabeled input space samples. For instance, a single input space sample is needed to check whether the model's output is positive or negative while two input space samples are needed to test the symmetry of the model w.r.t. its arguments. Constraints are rewritten in the standard equality and inequality form and $\rho^c_{j,k}$ is calculated as the constraint violation error. Coefficient $\beta$ is determined using the self-adaptive scheme.

- Regularization, $\mathcal{L}^r$ – this term drives the learning process towards a sparse neural network representing a concise analytic expression. Here, we adopt a smoothed $L_{0.5}$ regularization, $L^*_{0.5}$, as proposed in [21]. It exhibits several good properties. It is a trade-off between $L_0$ and $L_1$ regularization as it represents an optimization problem that can be solved using gradient descent, contrary to $L_0$, and at the same time, it enforces sparsity more strongly than $L_1$ while penalizing less the magnitude of the weights. Contrary to the original $L_{0.5}$ regularization, $L_{0.5}^*$ does not suffer from the singularity in the gradient as the weights converge to 0 and uses a piece-wise function to smooth out the function at small magnitudes. For a detailed setup of $L_{0.5}^*$, see [21]. The $\mathcal{L}^r$ term is calculated as the weighted $L_{0.5}^*$ value $\rho_k^r$ generated by all *active weights* $W_a \subseteq W_l$

$$\begin{aligned} \mathcal{L}^r &= \gamma \, \rho_k^r, \\ \rho_k^r &= \sum_{w \in W_a} L_{0.5}^*(w). \end{aligned} \quad (8)$$

A weight $w \in W_l$ is considered active if and only if its absolute value is larger than or equal to a user-defined threshold $\theta^a$ and it contributes to the NN output. The number of active weights, together with the number of active units, is used to report the NN model's complexity. The coefficient $\gamma$ is determined using the self-adaptive scheme. Note that $\rho_k^r$ values are not normalized by the mean of the historical values because we use just a single regularization type.

### D. SELF-ADAPTIVE LOSS TERMS WEIGHTING SCHEME

All three loss functions involve multiple terms, which leads to a multi-objective optimization problem. Besides the fact that the terms may be competing with each other (e.g., $\mathcal{L}^t$ vs. $\mathcal{L}^r$, the model's precision and complexity), they may also differ substantially in the scale of the values they attain. Since the final loss to be minimized is a weighted sum of the individual terms, undesired dominance of some terms may negatively influence the result. We, therefore, propose a self-adaptive method that adapts the coefficients $\alpha$, $\beta$, and $\gamma$ involved in the loss terms $\mathcal{L}^s$, $\mathcal{L}^c$, and $\mathcal{L}^r$ throughout the whole learning process in order to keep the desired ratios $r_{s/t} = \mathcal{L}^s : \mathcal{L}^t$, $r_{c/t} = \mathcal{L}^c : \mathcal{L}^t$, and $r_{r/t} = \mathcal{L}^r : \mathcal{L}^t$. It uses a sliding window strategy that works with the lists of values of $\mathcal{L}^t$, $\rho_{j,k}^s$, $\rho_{j,k}^c$, and $\rho_k^r$ observed in the last $N_w$ iterations of the learning process. The weights $\alpha$, $\beta$, and $\gamma$ are updated in each iteration according to

$$\begin{aligned} \alpha &= r_{s/t} \frac{\text{mean}(\mathbf{B}^t)}{\sum_{j \in T^s} \text{mean}(\{\frac{\rho_{j,k-i}^s}{h_{j,k-i}^s} : i = 0 \ldots N_w - 1\})}, \\ \beta &= r_{c/t} \frac{\text{mean}(\mathbf{B}^t)}{\sum_{j \in T^c} \text{mean}(\{\frac{\rho_{j,k-i}^c}{h_{j,k-i}^c} : i = 0 \ldots N_w - 1\})}, \\ \gamma &= r_{r/t} \frac{\text{mean}(\mathbf{B}^t)}{\text{mean}(\{\rho_{k-i}^r : i = 0 \ldots N_w - 1\})}, \end{aligned}$$

where $\mathbf{B}^t$ is the set of $N_w$ last $\mathcal{L}^t$ values. For $\alpha$ and $\beta$, if the denominator equals zero, the coefficient is set to one.

The coefficients are used to scale the respective loss term values, see equations (5), (7), and (8) so that the actual ratios stay close to the desired ratios. After the scaling is applied, i.e., the current values of $\mathcal{L}^s$, $\mathcal{L}^c$, and $\mathcal{L}^r$ have been calculated, it is further checked whether the obtained loss terms do not exceed the maximum value for which the actual ratio is less than or equal to the desired one. If this condition is violated, the respective loss term value is set to the value that implies the actual ratio is equal to the desired one, $r_{\_/t}$.

Note that only $\mathcal{L}^s$, $\mathcal{L}^c$, and $\mathcal{L}^r$ are scaled in each iteration. The $\mathcal{L}^t$ is a baseline relative to which the other terms are adjusted. Since the primary goal is to fit well the training data, each of $r_{s/t}$, $r_{c/t}$, and $r_{r/t}$ should be set to a value less than 1.

### E. FINAL MODEL SELECTION

During the gradient-based learning process, many NN models are generated. It is important to select the best one at the end, where the criteria are generally the model's complexity and its interpolation and extrapolation performance. In [18] and [19] as well as in [20] and [21], the final model selection method always builds on the fact that "a few" labeled samples from the unseen extrapolation domain are known, i.e., both the input variable values as well as the target value of the points are known. While the labeled extrapolation points are not used in the learning process, relying on specific information about the model's performance in the extrapolation domain for final model selection makes the approach dependent on that data. This means the extrapolation domain is no longer truly unseen. Additionally, these methods have limited practical use when extrapolation points cannot be measured on the system.

Here, we propose a final model section strategy that does not require labeled extrapolation points. Instead, it uses just the model's complexity, its validation RMSE, and the measures of its compliance with the prior knowledge and the singularity units' constraints. In practice, it is much easier to define these desired "high-level" model properties than to obtain particular measurements on the system. The proposed method uses an acceptance rule where one of the following two conditions must hold in order for the new *model* to be accepted as the best-so-far model $m^*$:

1) The *model*'s complexity is lower than the complexity of $m^*$.
2) The *model*'s complexity is equal to the complexity of $m^*$
    and
    for all $j \in T^s$, the *model*'s $\rho_{j,k}^s$ is smaller than or equal to that of $m^*$
    and
    for all $j \in T^c$, the *model*'s $\rho_{j,k}^c$ is smaller than or equal to that of $m^*$
    and
    the *model*'s validation RMSE is smaller than or equal to that of $m^*$.

Thus, the final $m^*$ is the least complex model found with the best values of all of $\rho_{j,k}^s$, $\rho_{j,k}^c$, and the validation RMSE objectives among the models of the same complexity. Note that the second condition can become true even if *model* is not strictly better than $m^*$ w.r.t. one or more performance indicators. This is to reduce the chance of getting stuck in some local optimum.

### F. EPOCH-WISE LEARNING PROCESS

The learning process, see Algorithm 1, is divided into three stages – initial, exploration-focus, and final stage.

The goal of the initial stage is to evolve the NN such that it exhibits at least partial capabilities 1) to fit the training data, 2) to satisfy the constraints imposed on the singularity units, and 3) to satisfy the constraints imposed on the desired model's properties. In the first half of this stage, the $\mathcal{L}_1$ loss function is optimized, while in the rest of this stage, $\mathcal{L}_2$ is optimized. The complexity of the NN does not matter at this stage. The NN then proceeds to the exploration-focus stage, where it is further trained epoch-wise. Each epoch starts with the exploration phase, lines 13–16, where all learnable weights $W_l$ are adjusted, followed by the focus phase, lines 17–22, where only active weights $W_a$ are further refined. The active weights are collected in the maskWeights() function at line 18 as the weights whose absolute value is above the threshold $\theta^a$. This is the only phase of the learning process where the $\mathcal{L}^r$ term is used to drive the search toward a simpler model. The purpose of the final stage is to fine-tune the active NN weights. All weights that become inactive at any iteration of this stage, line 25, are set to zero for the rest of the run, and only the active weights are updated in each learningStep() using $\mathcal{L}_2$. Thus, the complexity of the model can only decrease in this stage. Finally, the analytic expression represented by the best neural network instance found, $m^*$, is returned.

Depending on the loss function used, the respective weighting coefficients are updated after each learning step (e.g., lines 7–8). Besides the weight update, the learningStep() function updates several other objects. Firstly, it updates structures storing the history of the last $N_w$ values of the relevant loss terms. It also updates, when applicable, model $m^*$, see line 19 (and line 6 where $m^*$ is initialized for the first time), using the acceptance rule described in Section IV-E. Model $m^*$ serves as the seed for each epoch of the exploration-focus stage, line 11. Lastly, the function returns the current version of $model^*$, which is the best-so-far model with respect to the defined final model selection strategy; see Section IV-E.

The core of the learning process is the exploration-focus stage. This stage implements a restarted optimization strategy to avoid premature convergence to a potentially poor suboptimal model. It runs several epochs, each executing the exploration and focus phase one by one. At the beginning of each epoch, all learnable weights are set to the weights of the current $m^*$, line 11. Then, the maximum acceptable RMSE observed on the validation set $D_v$, $\theta^v$, is calculated, line 12. It is defined as $(1+\epsilon)$ times the mean validation RMSE over all models in $M^*$, which is a variable storing the final $m^*$ of the last $k$ epochs, see lines 9 and 23. The threshold $\theta^v$ is used in the acceptance rule that determines whether the new NN model updated within learningStep() will be accepted as $m^*$ of the current epoch. It gets accepted iff its complexity is not higher than the current $m^*$ complexity, and its validation RMSE is not higher than the threshold $\theta^v$. The parameter $\epsilon$ defines a tolerance margin, i.e., how much the current $m^*$ can be worse in terms of the validation RMSE than the mean of $k$ last $m^*$ models.

In the exploration phase, all weights in $W_l$ are optimized with respect to $\mathcal{L}_2$, line 14. After the exploration phase, the focus phase is carried out. Each iteration of this phase starts with extracting the set of active weights $W_a$, line 18, which are then optimized using $\mathcal{L}_3$. Once the focus phase has been completed, $M^*$ is updated using the current $m^*$.

---

**Algorithm 1** N4SR Algorithm

**Input:** Neural network with the set of learnable weights $W_l$
   $N_{init}, N_{final}$ ... number of iterations of the initial and final stage
   $N_w$ ... size of the loss term weights' adaptation window
   $N_e, N_f$ ... number of iterations of the exploration and focus phase
   $E$ ... number of the exploration-focus stage epochs
   $r_{s/t}, r_{c/t}, r_{r/t}$ ... desired loss terms' ratios
**Output:** Model in the form of an analytic expression represented by the final sparse network

---

1  init($W_l$)
  /* initial stage */
2  **for** $0 \leq i < N_{init}/2$ **do**
3    $B^t, B^s, model^* \leftarrow$ learningStep$_{\mathcal{L}_1}(W_l)$
4    $\alpha \leftarrow$ getTermWeight($B^t, B^s, r_{s/t}$)
5  **for** $0 \leq i < N_{init}/2$ **do**
6    $B^t, B^s, B^c, m^*, model^* \leftarrow$ learningStep$_{\mathcal{L}_2}(W_l)$
7    $\alpha \leftarrow$ getTermWeight($B^t, B^s, r_{s/t}$)
8    $\beta \leftarrow$ getTermWeight($B^t, B^c, r_{c/t}$)
9  $M^* \leftarrow m^*$
  /* exploration-focus stage */
10 **for** $0 \leq e < E$ **do**
11   $W_l \leftarrow$ getWeights($m^*$)
12   $\theta^v \leftarrow (1+\epsilon) \cdot$ getMeanValidRMSE($M^*$)
   /* exploration phase epochs */
13   **for** $0 \leq i < N_e$ **do**
14     $B^t, B^s, B^c, model^* \leftarrow$ learningStep$_{\mathcal{L}_2}(W_l)$
15     $\alpha \leftarrow$ getTermWeight($B^t, B^s, r_{s/t}$)
16     $\beta \leftarrow$ getTermWeight($B^t, B^c, r_{c/t}$)
   /* focus phase epochs */
17   **for** $0 \leq i < N_f$ **do**
18     $W_a \leftarrow$ maskWeights()
19     $B^t, B^s, B^c, B^r, m^*, model^* \leftarrow$
       learningStep$_{\mathcal{L}_3}(W_a)$
20     $\alpha \leftarrow$ getTermWeight($B^t, B^s, r_{s/t}$)
21     $\beta \leftarrow$ getTermWeight($B^t, B^c, r_{c/t}$)
22     $\gamma \leftarrow$ getTermWeight($B^t, B^r, r_{r/t}$)
23   $M^* \leftarrow$ update($M^*, m^*$)
  /* final stage */
24 **for** $0 \leq i < N_{final}$ **do**
25   $W_a \leftarrow$ maskWeights()
26   $B^t, B^s, B^c, model^* \leftarrow$ learningStep$_{\mathcal{L}_2}(W_a)$
27   $\alpha \leftarrow$ getTermWeight($B^t, B^s, r_{s/t}$)
28   $\beta \leftarrow$ getTermWeight($B^t, B^c, r_{c/t}$)
29 **return** getExpression($m^*$)

## V. EXPERIMENTS
### A. ALGORITHMS COMPARED
In this study, we experiment with three different methods. We compare multiple variants of N4SR, an alternative neural network-based method from the literature EQL$\div$, and a genetic programming-based algorithm mSNGP-LS.

The N4SR algorithms are divided into variants denoted as N4SR-WSCL with:

- Weighting W ∈ {A, S} – the adaptive (A) or static (S) loss terms weighting scheme. In the static variant, the weighting coefficients $\alpha$, $\beta$, and $\gamma$ are determined at the moment when they are used for the first time and stay constant for the rest of the run. In this way, it is at least partially ensured that all loss function terms will take on values in the relevant range. In particular, $\alpha$ is calculated at the very first iteration of the run using the initial $\mathcal{L}^t$ and $\rho_{j,0}^s$ values as $\alpha = r_{s/t} \cdot \mathcal{L}^t / \sum_{j \in T^s} \rho_{j,0}^s$. Similarly, $\beta$ is calculated at the first iteration of the second phase of the initial stage, line 8 of Algorithm 1, using the current $\mathcal{L}^t$ and $\rho_{j,k}^c$ values as $\beta = r_{c/t} \cdot \mathcal{L}^t / \sum_{j \in T^c} \rho_{j,k}^c$. Coefficient $\gamma$ is calculated at the first iteration of the first pass through the focus phase, line 22 of Algorithm 1, using the current $\mathcal{L}^t$ and $\rho_k^r$ values as $\gamma = r_{r/t} \cdot \mathcal{L}^t / \rho_k^r$. Then, the loss terms $\mathcal{L}^s$, $\mathcal{L}^c$, and $\mathcal{L}^r$ are calculated using (5), (7), and (8) while not applying the normalization in (5) and (7), so that $h_{j,k}^s = 1$, $h_{j,k}^c = 1$.
- Selection S ∈ {C, E} – the constraint satisfaction-based (C) final model selection rule, defined in Section IV-E, or an extrapolation-based (E) final model selection strategy. The extrapolation-based strategy is a modification of the one defined in Section IV-E such that only the *model*'s complexity and its RMSE observed on the extrapolation points are considered.
- Constraints C ∈ {Y, N} – denoting whether the constraints are used (Y) or not (N) within the learning process. The variant without constraints works according to Algorithm 1 with the modification that the loss functions $\mathcal{L}_2$ and $\mathcal{L}_3$ do not involve the $\mathcal{L}^c$ term. It also implies that this variant can work only with the extrapolation-based final model selection rule.
- Learning strategy L ∈ {E, S} – the epoch-wise (E) or a single-epoch (S) learning used within the exploration-focus stage. The single-epoch variant goes through a single epoch of the exploration-focus stage with $N_f$ set so that the total number of iterations spent in this stage is the same as for the epoch-wise variant.

Note that out of all the tested N4SR variants, the N4SR-ACYE one represents the proposed method and the other variants are used just for the ablation study purposes to analyze the effect of the four algorithm's components W, S, C, and L.

EQL$\div$ – the EQL algorithm with division units and the improved model selection method working with the extrapolation-validation dataset, proposed in [18]. We used the publicly available implementation.[3] Since, to our knowledge, no other NN-based SR method utilizes prior knowledge within the learning process, we can only compare with a "standard" SR technique, i.e., the one which uses the training data only. We are aware that such a comparison with EQL$\div$ is not entirely fair, so we use it just to illustrate that prior knowledge significantly improves the quality of the models generated.

mSNGP-LS – the multi-objective SNGP using the local search procedure to estimate coefficients of the evolved linear models, as proposed in [14].

### B. EVALUATION DATA SETS
The following data sets were used to evaluate models obtained with the compared algorithms:

- Extrapolation data set, $\bar{D}_e$ – contains data samples of the same form as $D_t$ sampled from the extrapolation domain. We use the extrapolation domain $\mathbb{D}_e$ to denote the parts of the problem's input space whose samples are either very sparsely present in $D_t$ and $D_v$ or are entirely omitted in these data sets. This data set is not used for learning – it is only used to select the final output model in the case of EQL$\div$ and the N4SR variants with S = E.
- Interpolation test data set, $D_i$ – contains data samples of the same form as $D_t$ sampled from $\mathbb{D}_t$ such that $D_i \cap (D_t \cup D_v) = \varnothing$. This data set is used to evaluate models' interpolation performance.
- Extrapolation test data set, $D_e$ – contains data samples of the same form as $D_t$ sampled from the extrapolation domain $\mathbb{D}_e$. This data set is used to evaluate models' extrapolation performance.

### C. TEST PROBLEMS
The proposed method was experimentally evaluated on the following four problems. We chose these problems since we possess detailed knowledge of the data and the desired properties of the models sought. Moreover, we can illustrate interesting application scenarios for these problems, such as using very sparse or unevenly distributed training data. Standard data-driven modeling approaches fail to generate physically plausible models when only such insufficient data sets are available.

#### 1) TurtleBot
The goal is to find a discrete-time model of a real physical system, the two-wheel TurtleBot 2 mobile robot (Figure 2). The robot's state is captured by the state vector $\mathbf{x} = (x_{pos}, y_{pos}, \phi)^\top$, with $x_{pos}$ and $y_{pos}$ the robot's position coordinates and $\phi$ the robot's heading. The control input is $\mathbf{u} = (v_f, v_a)^\top$, with $v_f$ and $v_a$ the desired forward and angular velocity, respectively. In this work, we model only the $x_{pos}$ component of the robot's motion model since 1) the $y_{pos}$ component is analogous and 2) it is more illustrative than the $\phi$ component as there are more types of prior knowledge

---

[3]https://github.com/martius-lab/EQL

defined for $x_{pos}$. The model has the form of the following nonlinear difference equation

$$x_{pos,k+1} = f^{x_{pos}}(x_{pos,k}, y_{pos,k}, \phi_k, v_{f,k}, v_{a,k}),$$

where $k$ denotes the discrete time step.

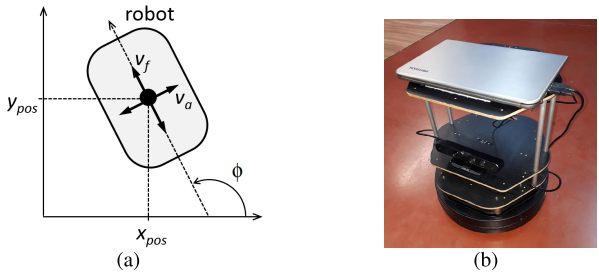

**FIGURE 2.** TurtleBot mobile robot. A schematic (a) and a photo of the system (b).

**Data sets**. We used the data sets introduced in [15], which were collected during the operation of the real robot. Five sequences of samples starting from the initial state $\mathbf{x_0} = (0, 0, 0)^\top$ were recorded with a sampling period $T_s = 0.2$ s. In each sequence, we steered the robot by random inputs drawn from the domain $v_f \in [0, 0.3]$ m·s$^{-1}$, $v_a \in [-1, 1]$ rad·s$^{-1}$. Of these five sequences, a randomly chosen one was used to create the training data set, another one was used as the validation data set, and the remaining three sequences were used as the test data sets.

**Prior knowledge**. We use the prior knowledge defined for the TurtleBot in [15]. All three prior knowledge types that the $x_{pos}$ variable should comply with are of the invariant type. This means that when the model for the $x_{pos}$ variable is evaluated on the relevant constraint sample, it should always output the next state equal to the current state. The following three types of prior knowledge about $x_{pos}$ were used:

a) Steady-state behavior: If the control inputs, $v_f$ and $v_a$, are set to zero, then the robot may change neither its position nor its heading. This is represented by the following equality constraint:

$$x_{pos} = f^{x_{pos}}(x_{pos}, y_{pos}, \phi, 0, 0).$$

b) Axis-parallel moves: If the robot moves parallel to the y-axis, then its $x_{pos}$ does not change. This is represented by the following equality constraints:

$$x_{pos} = f^{x_{pos}}(x_{pos}, y_{pos}, -\pi/2, v_f, 0),$$
$$x_{pos} = f^{x_{pos}}(x_{pos}, y_{pos}, \pi/2, v_f, 0).$$

c) Turning on the spot: If the forward velocity is zero, the robot may not change its position. This is represented by the following equality constraint:

$$x_{pos} = f^{x_{pos}}(x_{pos}, y_{pos}, \phi, 0, v_a).$$

The values of the state variables $x_{pos}$, $y_{pos}$, $\phi$, and of the control inputs $v_f$ and $v_a$ were randomly sampled within the same limits as for the training data. We generated 50 constraint samples for each prior knowledge type, so 150 samples in total.

#### 2) MAGNETIC MANIPULATION

The magnetic manipulation system, magman, consists of an iron ball moving on a rail and an electromagnet placed at a fixed position under the rail (Figure 3). The goal is to find a nonlinear model of the magnetic force, $f(x)$, affecting the ball, as a function of the horizontal distance, $x$, between the iron ball and the electromagnet, given a constant current $i$ through the coil. We use data measured on the real system [38] to train and evaluate the models. To further evaluate the extrapolation performance of the models, we use the data sampled from an empirical model $\tilde{f}(x) = -ic_1 x/(x^2 + c_2)^3$ proposed in [39], where the parameters $c_1$ and $c_2$ were found empirically for the given system. This model serves as the *reference model*, see Figure 4.

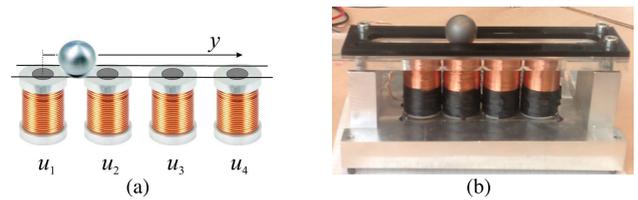

**FIGURE 3.** Magman: A schematic (a) and a photo of the system (b).

**Data sets**. The whole data set of 858 samples measured on the real system was split into two sets in the ratio of 7:3. The larger one was further split into the $D_t$ and $D_v$ sets so that $|D_t| = 400$ and $|D_v| = 201$. The smaller one was used as the test interpolation data set $D_i$. The operating region of magman is the interval $-0.075$ m $\leq x \leq 0.075$ m. However, only its small part, $\mathbb{D}_t = [-0.027, 0.027]$ m, is covered by the data, see Figure 4. A proper form of the model outside the sampled interval is induced by the constraints imposed on the model, see below. Additionally, two data sets, $\bar{D}_e$ and $D_e$, with samples from the extrapolation domain $\mathbb{D}_e = [-0.075, -0.027] \cup [0.027, 0.075]$ m were generated with the sizes $|\bar{D}_e| = 40$ and $|D_e| = 200$. The target values of $\bar{D}_e$ and $D_e$ samples were computed using the reference model.

**Prior knowledge**. Five types of prior knowledge were defined. The function sought is odd: it returns positive values on the interval $[-0.075, 0]$ m and negative ones on the interval $[0, 0.075]$ m. It is monotonically increasing on the intervals $[-0.075, -0.008]$ m and $[0.008, 0.075]$ m and monotonically decreasing on the interval $[-0.008, 0.008]$ m. Finally, the function passes through the origin, i.e., $f(0) = 0$, and quickly decays to zero as $x$ tends to infinity, which is represented by the constraints $f(-0.075) = 10^{-3}$ and $f(0.075) = -10^{-3}$, respectively. The constraint set contains 50 samples for the positive values, negative values, increasing monotonicity, and decreasing monotonicity plus 3 samples for the desired exact values, resulting in the total of 203 constraint samples.

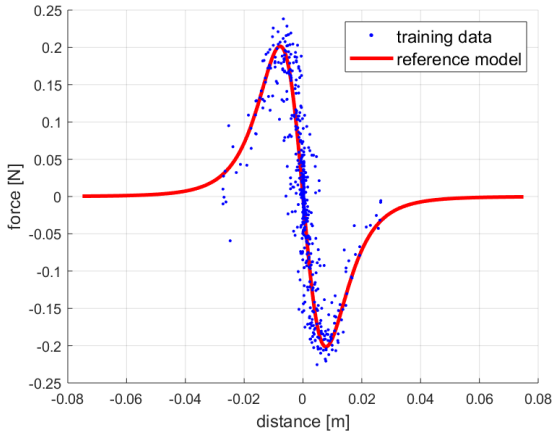

**FIGURE 4.** Training data and the reference model for the `magman` problem.

### 3) RESISTORS

This problem was proposed in [8] to test SR method based on genetic programming with formal constraints. Originally, it considers a sparse set of noisy samples derived using the equivalent resistance of two resistors in parallel, $r = r_1 r_2/(r_1 + r_2)$, used here as the true system. The goal is to find such a model $f(r_1, r_2)$ that fits the training data and exhibits the same properties as the *reference model*, see below.

**Data sets**. Here, we use two variants of the data set: one with only 10 samples and the other one with 500 samples. The values of $r_1$ and $r_2$ are sampled uniformly from the interval $[20.0001]\,\Omega$. The target values are corrupted by random noise with the normal distribution $\mathcal{N}(0, 0.05\,\sigma_y)$, where $\sigma_y$ is a standard deviation of the original output values. When using the large set, it is split into $D_t$ and $D_v$ in the ratio of 7:3. When the smaller one is used, eight samples go to $D_t$, and the remaining two samples are in $D_v$. The interpolation test set $D_i$ is sampled from the training domain as well. Two data sets are sampled from the extrapolation domain $\mathbb{D}_e = [20, 40]^2$, $\bar{D}_e$ with 40 samples and $D_e$ with 500 samples.

**Prior knowledge**. We used the following three prior knowledge types as defined in [8] and used also in [14]:

- symmetry with respect to the arguments:
  $f(r_1, r_2) = f(r_2, r_1)$,
- domain-specific constraint:
  $r_1 = r_2 \implies f(r_1, r_2) = \frac{r_1}{2}$,
- domain-specific constraint:
  $f(r_1, r_2) \leq r_1, f(r_1, r_2) \leq r_2$.

The constraint set contains a total of 150 samples, 50 for each constraint.

### 4) ANTI-LOCK BRAKING SYSTEM – MAGIC FORMULA

This problem considers the tire-road interaction model, particularly the longitudinal force $F(\kappa)$ as a function of the wheel slip $\kappa$. The force $F(\kappa)$ is described by the 'magic' formula of the following form

$$F(\kappa) = m\,g\,d\,\sin(c \arctan(b(1-e)\kappa + e \arctan(b\kappa))), \quad (9)$$

where $b$, $c$, $d$ and $e$ are road surface-specific constants. The magic formula is an empirical model commonly used to simulate steady-state tire forces and moments. By adjusting the function coefficients, the same special function can be used to describe longitudinal and lateral forces (sine function) and self-aligning moment (cosine function). Here, we consider the *reference model* used in [40] with $m = 407.75$ kg, $g = 9.81\,\mathrm{m\cdot s^{-1}}$, and the slip force parameters $(b, c, d, e) = (55.56, 1.35, 0.4, 0.52)$, see Figure 5, which correspond to a wet asphalt for a water level of 3 mm [41].

**Data sets**. A data set of 110 samples was generated using the reference model (9). The whole set was divided into $D_t$ and $D_v$ in the ratio of 4:1. The data are intentionally sampled unevenly. The steep left and the right flat region of $\kappa$ in the interval $[0, 0.02]$ and $[0.2, 0.99]$, representing the interpolation domain $\mathbb{D}_i$, are densely sampled with 50 samples each. Target values of these samples are disturbed with a noise randomly generated with a normal distribution $\mathcal{N}(0, 0.0025)$. Contrary, the peak of the function with $\kappa$ values in the interval $[0.03, 0.1]$ is covered very sparsely in the data with only 10 samples. Moreover, a larger noise drawn from $\mathcal{N}(0, 0.005)$ is added to the target values. This corresponds to the fact that in reality, the system is unstable around the peak and it is hard to collect precise data there. Thus, the peak represents the extrapolation domain $\mathbb{D}_e = [0.03, 0.1]$ in the sense that it is rather poorly defined by the data. Again, the data deficiency is compensated in our method by the use of prior knowledge, see below. Additionally, three data sets for models' performance evaluation were used: $D_i$ of size 200 sampled from $\mathbb{D}_i$ and data sets $\bar{D}_e$ and $D_e$ of sizes 40 and 100, respectively, sampled from $\mathbb{D}_e$. The target values of the samples were generated as the noiseless output of the reference model.

**Prior knowledge**. Three types of prior knowledge were defined for this problem reflecting the key properties of the model sought. The model should return zero for $\kappa = 0$. Further, in the right part of $\mathbb{D}_i$, the model is monotonically decreasing and in the limit approached from above a constant value. So, its second derivative in this region should be positive. The last property of the model is that it has a single maximum located within $\mathbb{D}_e$. This can be described by a constraint that enforces the model to be concave down everywhere in $\mathbb{D}_e$. The constraint set contains a total of 150 constraint samples, 50 samples for each constraint.

We briefly illustrate the implementation of constraints for a monotonically decreasing function with a positive second derivative. As described in Section IV-A, a set of $N_j$ constraint samples is generated for each constraint $j$. Here, the samples have the form $(\kappa_l, \kappa_c, \kappa_r)$, where $\kappa_c$ is randomly sampled in $\mathbb{D}_e$ and $\kappa_l = \kappa_c - \delta$, $\kappa_r = \kappa_c + \delta$, and $\delta = 0.001$. For each such constraint sample $k$, the error value $e_k$ is calculated as

$$e_k = \max((f(\kappa_r) - f(\kappa_c)), 0) + \max((f(\kappa_c) - f(\kappa_l)), 0)$$
$$+ \max((f(\kappa_c) - f(\kappa_r)) - (f(\kappa_l) - f(\kappa_c)), 0),$$

where the first line represents contributions for a non-decreasing property observed on the given constraint data triple ($\kappa_l$, $\kappa_c$, $\kappa_r$) and the second line represents a penalty for a negative second derivative observed on the constraint sample. Then, the corresponding $\rho_{j,k}^c$ is calculated as the root-mean-square error over all $e_k$ observed for the constraint.

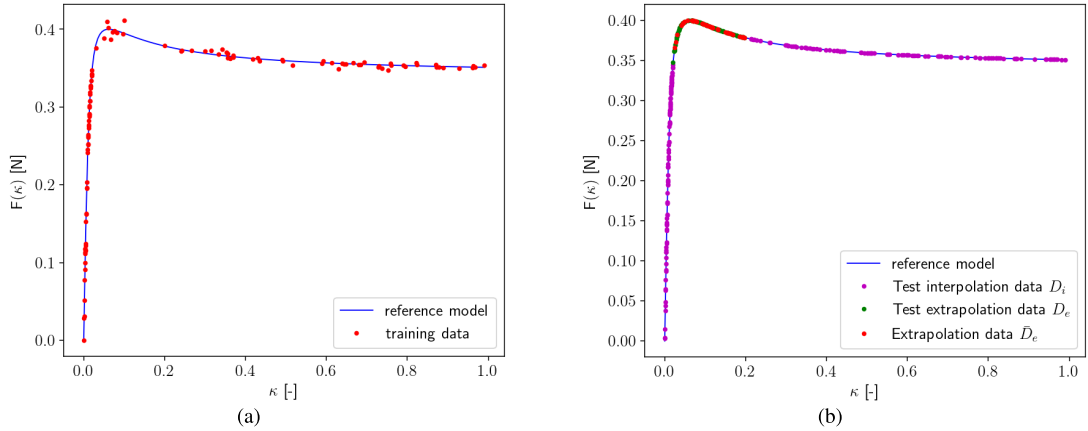

**FIGURE 5.** Magic formula reference model and the data sets. (a) Training and validation data. (b) Interpolation test data and extrapolation data.

### D. EXPERIMENT SET UP
#### 1) NETWORK ARCHITECTURE
We used a network architecture with three hidden layers, denoted as the 'general' architecture. The first two hidden layers contain four elementary functions {sin, tanh, ident, ∗}, each with two copies. The third hidden layer contains, in addition to that, one division unit. The output layer contains a single identity unit ident. The ident unit calculates a weighted sum of its inputs so it can realize both the addition and subtraction units. The same architecture was used for all test problems but the `magic` one. There we used a function set with arctan instead of tanh in order to make the set-up consistent with the magic formula reference model (9). Moreover, for experiments with the `resistors` problem, we used an additional architecture with a limited function set {ident, ∗, /}, denoted as the 'informed' architecture, that corresponds to the one used in [14]. Its name reflects the fact that we know the set of elementary functions needed to compose the correct formula. The first two hidden layers contain 4 copies of the multiplication and ident unit each. The third hidden layer contains 3 copies of each of those two unit types. The number of units was chosen so that the number of learnable weights of the two architectures was as close as possible. Thus, the general and the informed architecture comprise 396 and 403 learnable weights in total, respectively. For the other problems using the general architecture, the number of learnable weights was 495 (`turtlebot`) and 363 (`magman` and `magic`).

#### 2) ALGORITHMS' CONFIGURATION
The algorithms were tested with the following parameter setting:

- `N4SR`: $N_{init}$ = 2000, $N_e$ = 20, $N_f$ = 980 and $E$ = 87 for epoch-wise variants, $N_f$ = 86980 and $E$ = 1 for single epoch variants, $N_{final}$ = 1000, $N_w$ = 10 the parameters are chosen so that the total number of iterations is always $T$ = 90000, $r_{s/t}$ = 0.5, $r_{c/t}$ = 0.5, $r_{r/t}$ = 0.5, $\theta_a$ = 0.0001, $\theta_j^s$ = 0.0001, $\epsilon$ = 0.5.
- `EQL`$^{\div}$: We adopted the configuration used in [18] with the exception that the total number of iterations $T$ was set to 90000. The network's topology was set the same as in the corresponding `N4SR` experiments, except that the division unit is not in the third hidden layer. Instead, it is the only unit of the output layer, which is a design feature of `EQL`$^{\div}$.
- `mSNGP-LS`: We adopted the configuration used in [14] with a few modifications. The set of elementary functions is set to comply with the set of elementary functions used in the `N4SR` networks for the given problem. The maximum model's complexity is bounded from above by two parameters, the maximum number of features $n_f$ = 5 and the maximum feature depth $\delta$ = 3, used with the same values for all test problems. Note the maximum feature depth is equal to the number of hidden layers of the `N4SR` networks. Thus, when the features are aggregated in the final model, its maximum possible depth is the same as the maximum depth of the `N4SR` models.

Note that all tested methods used the same parameter setting on all test problems. No parameter tuning was carried out.

#### 3) EXPERIMENTS EVALUATION
Fifty independent runs were carried out for each tested pair of the method and the training data set. The best model is returned at the end of each run according to the model selection strategy used. For the `mSNGP-LS`, we adopt the selection method proposed in [15]. It uses two performance metrics — the RMSE calculated on the validation data set $D_v$ and the constraint violation error observed on the validation constraint data set. The validation constraint data set is generated the same way as the constraint samples used in N4SR methods, and the constraint violation error is calculated as the sum of $\rho_{j,k}^c$ values for all constraints in $T^c$. Then, the mSNGP-LS method chooses among all models in the last population of the run the model with the best validation RMSE out of all models with the constraint violation value less than the population's median.

**TABLE 1.** Results obtained with EQL$^{\div}$, mSNGP-LS, and N4SR methods with the general architecture on the turtlebot problem. The complexity is given as the number of active links / number of active units. The complexity and the RMSE values are medians over 50 runs. The *p*-values relate to the statistical tests calculated on the *RMSE$_{sum}$* values.

| Method | complexity | $N_{nt}$ | $RMSE_{valid}$ | $RMSE_{test1}$ | $RMSE_{test2}$ | $RMSE_{test3}$ | $RMSE_{sum}$ | *p*-value |
|---|---|---|---|---|---|---|---|---|
| **N4SR-ACYE** | 15 / 4 | 49 | 0.21 | 0.49 | 0.53 | 0.44 | 1.43 | – |
| N4SR-ACYS | 19 / 6 | 50 | 0.28 | 0.34 | 0.53 | 0.34 | 1.15 | $6.85 \cdot 10^{-1}$ |
| N4SR-AEYE | 12 / 4 | 50 | 0.17 | 0.44 | 0.41 | 0.41 | 1.23 | $6.69 \cdot 10^{-1}$ |
| N4SR-AEYS | 21 / 6 | 50 | 0.20 | 0.37 | 0.43 | 0.34 | 1.05 | $3.34 \cdot 10^{-1}$ |
| N4SR-AENE | 16 / 4 | 50 | 5.97 | 2.40 | 1.01 | 2.30 | 5.52 | $\mathbf{6.24 \cdot 10^{-13}}$ |
| N4SR-SCYE | 22 / 4 | 50 | 2.30 | 2.05 | 1.37 | 1.84 | 5.25 | $\mathbf{4.25 \cdot 10^{-12}}$ |
| N4SR-SCYS | 27 / 6 | 50 | 2.28 | 1.81 | 1.26 | 1.78 | 5.31 | $\mathbf{6.29 \cdot 10^{-11}}$ |
| N4SR-SENE | 15 / 4 | 50 | 4.30 | 2.61 | 2.33 | 2.52 | 7.46 | $\mathbf{4.54 \cdot 10^{-12}}$ |
| EQL$^{\div}$ | 30 / 18 | 50 | 1.73 | 2.61 | 1.97 | 2.26 | 8.68 | $\mathbf{1.44 \cdot 10^{-17}}$ |
| mSNGP-LS | NA | 50 | 0.03 | 0.11 | 0.17 | 0.15 | **0.42** | $\mathbf{8.72 \cdot 10^{-11}}$ |

On the turtlebot problem, the models' performance is presented using a simulation RMSE, which is calculated as the root-mean-square-error between $x_{pos,k+1}$ and $\hat{x}_{pos,k+1}$ for all points $k = 1 \ldots |D| - 1$ in the data sequence $D$, where $\hat{x}_{pos,k+1}$ is the value predicted by the model according to $\hat{x}_{pos,k+1} = f^{x_{pos}}(\hat{x}_{pos,k}, y_{pos,k}, \phi_k, v_{f,k}, v_{a,k})$.

Finally, the median values of the following performance measures over the fifty best-of-run models are presented:

- complexity – model complexity defined as the number of active weights and active units, respectively.
- $RMSE_{int}$, $RMSE_{ext}$, $RMSE_{int+ext}$ – RMSE calculated on test data $D_i$, $D_e$, and $D_i \cup D_e$, respectively.
- $RMSE_{valid}$, $RMSE_{test1}$, $RMSE_{test2}$, $RMSE_{test3}$ – simulation RMSE values calculated on the turtlebot problem, and the $RMSE_{sum}$ value calculated as the sum of all $RMSE_{test}$ values.
- $N_{nt}$ – the number of runs in which the method yields a nontrivial model, i.e., a model with more than one active link.

To assess the statistical significance of the differences among the methods, we analyze them pair-wise, N4SR-ACYE vs every other method and N4SR variant, using the Wilcoxon rank-sum test. We use the significance level of 5 % to reject the null hypothesis that the compared sets of the respective performance measure values are sampled from continuous distributions with equal medians. All tables present the *p*-value for each algorithm compared (highlighted in bold when it is less than 0.05).

### E. RESULTS

In this section, results are presented for all of the compared methods, accompanied by examples of models produced by the N4SR variants. In the tables, the **N4SR–ACYE** is highlighted in bold as this is the N4SR variant that implements the proposed method.

#### 1) TurtleBot

Table 1 shows results obtained on the turtlebot problem. We can see that N4SR-ACYE significantly outperforms all static variants on the test sequences. This is indicated by its smaller median $RMSE_{sum}$ value and very small *p*-values. The epoch-wise learning strategy performs comparably to the single-epoch one in terms of $RMSE_{sum}$; see N4SR-ACYE to N4SR-ACYS, N4SR-AEYE to N4SR-AEYS, and N4SR-SCYE to N4SR-SCYS. Nevertheless, the epoch-wise strategy generates significantly simpler models in terms of the number of active weights and active units.

Interestingly, the proposed constraint satisfaction-based final model selection method works comparably to the extrapolation-based one; see N4SR-ACYE vs. N4SR-AEYE and N4SR-ACYS vs. N4SR-AEYS. This is an important observation that demonstrates the ability of the method to reliably identify the final model even when no extrapolation domain samples are provided. Further, we observe a clear benefit of using prior knowledge for learning. When no prior knowledge is used, poor models are produced, see N4SR-SENE, N4SR-AENE, and the EQL$^{\div}$ models. Finally, the best-performing method on this problem is the mSNGP-LS. We discuss a possible reason for what causes the difference between N4SR and mSNGP-LS in Section V-F.

Figure 6 shows examples of trajectories generated with selected models. Particularly, the median and the best models w.r.t. $RMSE_{sum}$ produced by N4SR-ACYE (Figure 6a–c) and N4SR-AENE (Figure 6d–f) are presented. These plots clearly illustrate the benefits of using prior knowledge. One can see that both the median and the best model produced by N4SR-ACYE generate trajectories that accurately imitate the shape of the ground truth one. Moreover, the trajectories generated with the best model have a minimal offset from the ground truth one along the whole trajectory. On the contrary, trajectories generated with the models produced by N4SR-AENE exhibit larger discrepancies in terms of both the shape and the offset.

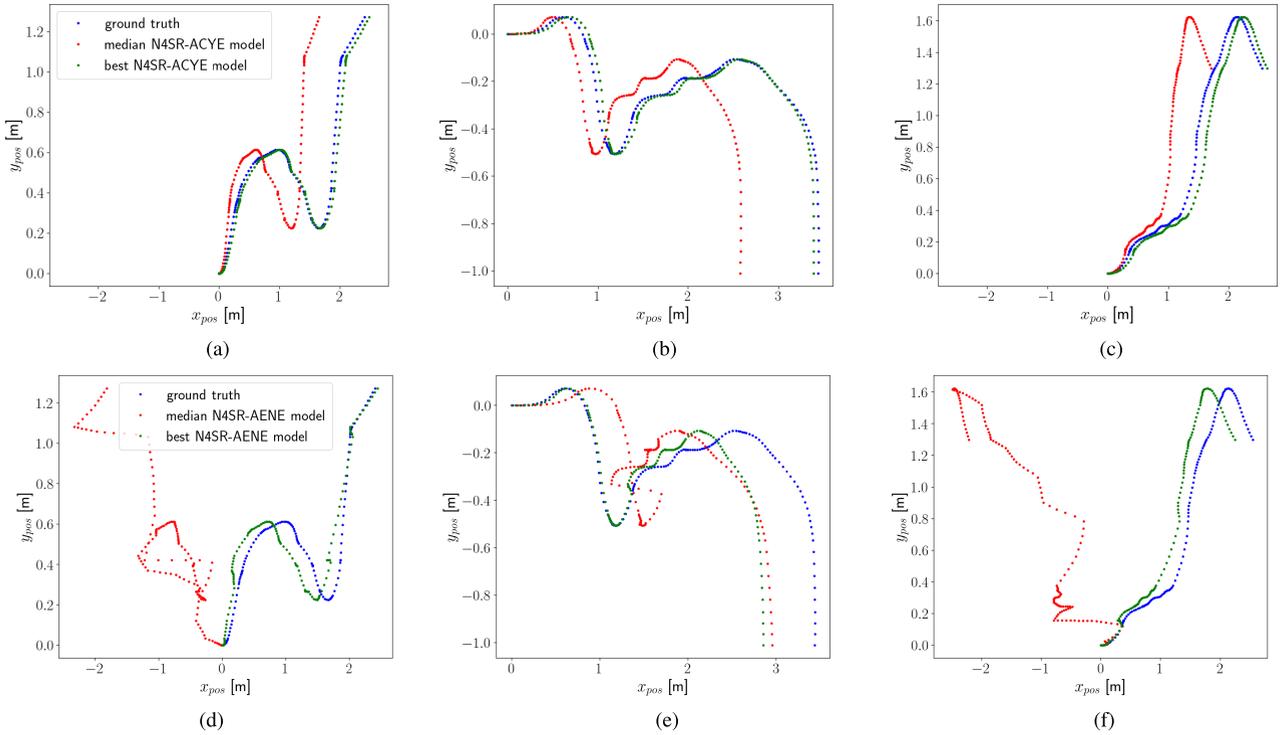

**FIGURE 6.** Examples of the `turtlebot` simulation trajectories generated using the $f^{x_{pos}}(\cdot)$ models obtained with `N4SR-ACYE` (a)-(c) and `N4SR-AENE` (d)-(f), respectively, on the three test data sets. Out of all best-of-run models collected over all runs for each experiment, a trajectory of the median model w.r.t. $RMSE_{sum}$ (shown in red) and a trajectory of the model with the best $RMSE_{sum}$ value (shown in green) are presented. The reference ground truth trajectory is shown in blue.

The raw analytic expression represented by the best `N4SR-ACYE` model is

$$x_{pos,k+1} = -0.3822\,((0.5605\,\sin(0.9838\,\phi_k + 1.5337))\\(-0.7903\,v_{f,k})) + 0.999986\,x_{pos,k},$$

that can be further simplified to

$$x_{pos,k+1} = 0.1693\,v_{f,k}\,\sin(0.9838\,\phi_k + 1.5337)\\+ 0.999986\,x_{pos,k}.$$

We can see that the best `N4SR-AENE` model, represented by the following simplified analytic expression, is much more complex:

$$x_{pos,k+1} = 0.170 - 0.088\,v_{f,k} + 1.0003\,x_{pos,k} + 0.169\\(0.555\,\sin(-0.140\,\phi_k^2 + 0.871\,\phi_k\,v_{f,k} + 0.087\,\phi_k\\+ 2.587\,v_{f,k} + 0.281) + 0.204)\,(1.309\,v_{f,k} + 0.053\,y_{pos,k}\\- 0.223(0.535\phi_k + 1.588)(0.271\phi_k - 1.691 v_{f,k} + 0.489)\\+ 0.096\,\sin(-0.140\,\phi_k^2 + 0.871\,\phi_k\,v_{f,k} + 0.087\,\phi_k\\+ 2.587\,v_{f,k} + 0.281)).$$

#### 2) MAGNETIC MANIPULATION

The results obtained on the `magman` problem are shown in Table 2. We can see that `N4SR-ACYE` and `N4SR-ACYS` produce models that have the best performance on the extrapolation data as its $RMSE_{ext}$ is, by order of magnitude, better than the `EQL`$^\div$, `mSNGP-LS`, and `N4SR` variants using the static weighting scheme. Only models obtained with `N4SR-AEYE` and `N4SR-AEYS` exhibit better extrapolation performance. But this can be attributed to the fact that these variants already use certain knowledge of the models' extrapolation performance when selecting the final model of each run. `N4SR-ACYE` performs the best, even better than `mSNGP-LS`, in terms of the overall $RMSE_{int+ext}$ metric. Again, the constraint satisfaction-based final model selection performs equally to the extrapolation-based one.

All adaptive `N4SR` variants but `N4SR-AENE` are highly vulnerable to collapsing to a trivial model with only one active weight, particularly the one of the output unit's bias link. Only about 30 % of the runs end up with nontrivial models. `N4SR-AENE` finds nontrivial models in 38 runs, but these fit well only in the interpolation domain. Outside the interpolation domain, the models go wild since the method cannot use any helpful information to direct the search toward better models. Contrary to the adaptive `N4SR` variants, the static ones generate nontrivial models in 90 % of runs. However, the models obtained with the static methods perform significantly worse in terms of $RMSE_{int+ext}$ than the ones generated by `N4SR-ACYE`.

Figure 7 shows models generated by `N4SR-ACYE`, `N4SR-SCYE`, `EQL`$^\div$, and `mSNGP-LS`. Again, the median and the best models w.r.t. *RMSE*$_{int+ext}$ over the set of nontrivial models are presented. One can see that neither of the `N4SR-SCYE`, `EQL`$^{\div}$, and `mSNGP-LS` methods can reliably produce nontrivial models that obey the increasing monotonicity constraint. In particular, `N4SR-SCYE` and `mSNGP-LS` succeeded only in 6 and 3 out of 50 runs, and `EQL`$^{\div}$ failed to generate such a model in all runs. Contrary to that, all nontrivial models produced by `N4SR-ACYE` comply with this monotonicity constraint. Also interesting is the fact that both `N4SR` variants are doing very well in terms of the property that the model goes through the origin while the `EQL`$^{\div}$ and `mSNGP-LS` have certain offset at $x = 0$ m.

**TABLE 2.** Results obtained with `EQL`$^{\div}$, `mSNGP-LS`, and the `N4SR` methods with the general architecture on the `magman` problem. The complexity is given as the number of active links / number of active units. The complexity and the RMSE values are medians over 50 runs. The *p*-values relate to the statistical tests calculated on the *RMSE*$_{int+ext}$ values.

| Method | complexity | $N_{nt}$ | $RMSE_{int}$ | $RMSE_{ext}$ | $RMSE_{int+ext}$ | *p*-value |
|---|---|---|---|---|---|---|
| **N4SR-ACYE** | 9 / 5 | 16 | $5.81 \cdot 10^{-2}$ | $5.27 \cdot 10^{-3}$ | $4.37 \cdot 10^{-2}$ | – |
| N4SR-ACYS | 8 / 5 | 15 | $5.81 \cdot 10^{-2}$ | $6.64 \cdot 10^{-3}$ | $4.38 \cdot 10^{-2}$ | $3.3 \cdot 10^{-1}$ |
| N4SR-AEYE | 9 / 5 | 19 | $5.83 \cdot 10^{-2}$ | $1.42 \cdot 10^{-3}$ | $4.38 \cdot 10^{-2}$ | $4.8 \cdot 10^{-1}$ |
| N4SR-AEYS | 7.5 / 5 | 18 | $5.87 \cdot 10^{-2}$ | $6.26 \cdot 10^{-4}$ | $4.41 \cdot 10^{-2}$ | $\mathbf{1.6 \cdot 10^{-2}}$ |
| N4SR-AENE | 7.5 / 5 | 38 | $5.70 \cdot 10^{-2}$ | $2.46 \cdot 10^{-1}$ | $1.12 \cdot 10^{-1}$ | $\mathbf{4.1 \cdot 10^{-9}}$ |
| N4SR-SCYE | 11 / 6 | 46 | $5.64 \cdot 10^{-2}$ | $6.01 \cdot 10^{-2}$ | $5.68 \cdot 10^{-2}$ | $\mathbf{2.21 \cdot 10^{-7}}$ |
| N4SR-SCYS | 10.5 / 6 | 46 | $5.64 \cdot 10^{-2}$ | $5.46 \cdot 10^{-2}$ | $5.57 \cdot 10^{-2}$ | $\mathbf{4.73 \cdot 10^{-7}}$ |
| N4SR-SENE | 8 / 5 | 42 | $5.70 \cdot 10^{-2}$ | $1.75 \cdot 10^{-1}$ | $1.23 \cdot 10^{-1}$ | $\mathbf{3.3 \cdot 10^{-9}}$ |
| EQL$^{\div}$ | 30 / 18 | 50 | $5.60 \cdot 10^{-2}$ | $6.12 \cdot 10^{-2}$ | $5.86 \cdot 10^{-2}$ | $\mathbf{2.3 \cdot 10^{-8}}$ |
| mSNGP-LS | NA | 50 | $5.59 \cdot 10^{-2}$ | $4.61 \cdot 10^{-2}$ | $5.18 \cdot 10^{-2}$ | $\mathbf{4.6 \cdot 10^{-4}}$ |

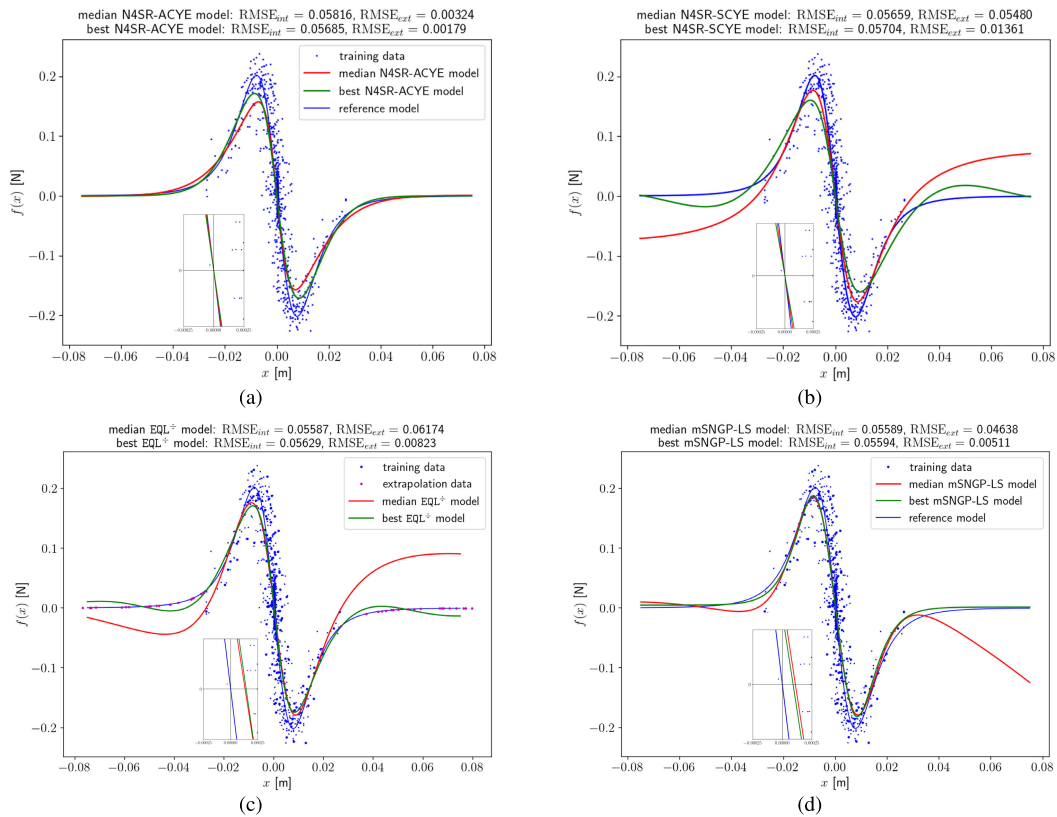

**FIGURE 7.** Examples of models generated for the `magman` problem with `N4SR-ACYE` (a), `N4SR-SCYE` (b), `EQL`$^{\div}$ (c), and `mSNGP-LS` (d) method. Out of all best-of-run models collected over all the runs, the median model w.r.t. *RMSE*$_{int+ext}$ (shown in red) and the model with the best *RMSE*$_{int+ext}$ value (shown in green) are presented.

For illustration, we show the best `N4SR-ACYE` model

$$f(x) = -0.429 \sin(1.971 \tanh(4.294\,x)$$
$$+ 5.33 \tanh(8.494 \tanh(4.294\,x)))$$
$$- 2.027 \tanh(4.701 \tanh(8.494 \tanh(4.294\,x)))$$
$$+ 1.607 \tanh(8.494 \tanh(4.294\,x))$$
$$+ 0.873 \tanh(4.294\,x)$$

and the best `N4SR-SCYE` model

$$f(x) = -1.731 \sin(85.834\,x)/(3354.189\,x^2 + 0.484).$$

**TABLE 3.** Results obtained with `EQL÷`, `mSNGP-LS`, and `N4SR` methods with the general architecture on the `resistors` problem. The complexity is given as the number of active links / number of active units. The complexity and the RMSE values are medians over 50 runs. The p-values relate to the statistical tests calculated on the $RMSE_{int+ext}$ values.

| $|D_t \cup D_v|$ | Method | Complexity | $N_{nt}$ | $RMSE_{int}$ | $RMSE_{ext}$ | $RMSE_{int+ext}$ | p-value |
|---|---|---|---|---|---|---|---|
| 500 | **N4SR−ACYE** | 10 / 3 | 49 | 0.035 | 0.056 | 0.047 | − |
| 500 | N4SR-ACYS | 10 / 4 | 49 | 0.210 | 0.249 | 0.589 | $2.69 \cdot 10^{-2}$ |
| 500 | N4SR-AEYE | 11 / 3 | 50 | 0.061 | 0.076 | 0.069 | $6.44 \cdot 10^{-1}$ |
| 500 | N4SR-AEYS | 9 / 4 | 47 | 0.287 | 0.887 | 0.702 | $2.43 \cdot 10^{-3}$ |
| 500 | N4SR-AENE | 36.5 / 9 | 50 | 0.085 | 0.702 | 0.502 | $5.92 \cdot 10^{-6}$ |
| 500 | N4SR-SCYE | 46 / 11 | 50 | 0.083 | 0.345 | 0.251 | $7.42 \cdot 10^{-2}$ |
| 500 | N4SR-SCYS | 53 / 13 | 50 | 0.077 | 0.305 | 0.225 | $1.38 \cdot 10^{-1}$ |
| 500 | N4SR-SENE | 45 / 11 | 50 | 0.092 | 0.811 | 0.578 | $3.08 \cdot 10^{-5}$ |
| 500 | EQL÷ | 30 / 18 | 50 | 0.550 | 8.460 | 6.010 | $1.96 \cdot 10^{-17}$ |
| 500 | mSNGP-LS | NA | 50 | 0.023 | 0.032 | **0.029** | $4.99 \cdot 10^{-1}$ |
| 10 | **N4SR−ACYE** | 25 / 7 | 50 | 0.377 | 1.235 | 0.927 | − |
| 10 | N4SR-ACYS | 25 / 8 | 50 | 0.601 | 1.561 | 1.156 | $2.1 \cdot 10^{-1}$ |
| 10 | N4SR-AEYE | 22 / 6 | 50 | 0.623 | 1.157 | 0.948 | $7.88 \cdot 10^{-1}$ |
| 10 | N4SR-AEYS | 21 / 7 | 50 | 0.657 | 1.749 | 1.311 | $2.41 \cdot 10^{-1}$ |
| 10 | N4SR-AENE | 24 / 7 | 50 | 1.040 | 2.370 | 1.850 | $1.54 \cdot 10^{-13}$ |
| 10 | N4SR-SCYE | 38 / 9 | 50 | 0.623 | 0.822 | 0.738 | $4.47 \cdot 10^{-1}$ |
| 10 | N4SR-SCYS | 39 / 9 | 50 | 0.589 | 0.782 | 0.700 | $3.39 \cdot 10^{-1}$ |
| 10 | N4SR-SENE | 29 / 8 | 50 | 1.050 | 2.443 | 1.885 | $1.76 \cdot 10^{-12}$ |
| 10 | EQL÷ | 30 / 18 | 50 | 18.820 | 17.420 | 22.810 | $9.26 \cdot 10^{-18}$ |
| 10 | mSNGP-LS | NA | 50 | 0.008 | 0.009 | **0.008** | $9.03 \cdot 10^{-8}$ |

**TABLE 4.** Results obtained with `EQL÷`, `mSNGP-LS`, and `N4SR` methods with the informed architecture on the `resistors` problem. The complexity is given as the number of active links / number of active units. The complexity and the RMSE values are medians over 50 runs. The p-values relate to the statistical tests calculated on the $RMSE_{int+ext}$ values.

| $|D_t \cup D_v|$ | Method | complexity | $N_{nt}$ | $RMSE_{int}$ | $RMSE_{ext}$ | $RMSE_{int+ext}$ | p-value |
|---|---|---|---|---|---|---|---|
| 500 | **N4SR−ACYE** | 10 / 3 | 49 | 0.007 | 0.011 | **0.009** | − |
| 500 | N4SR-ACYS | 8 / 3 | 49 | 0.008 | 0.013 | 0.011 | $6.16 \cdot 10^{-1}$ |
| 500 | N4SR-AEYE | 10 / 3 | 49 | 0.009 | 0.015 | 0.012 | $5.37 \cdot 10^{-2}$ |
| 500 | N4SR-AEYS | 8 / 3 | 50 | 0.009 | 0.016 | 0.014 | $2.51 \cdot 10^{-1}$ |
| 500 | N4SR-AENE | 12.5 / 4 | 50 | 0.050 | 0.040 | 0.063 | $7.69 \cdot 10^{-13}$ |
| 500 | N4SR-SCYE | 15 / 3 | 50 | 0.039 | 0.071 | 0.056 | $1.82 \cdot 10^{-12}$ |
| 500 | N4SR-SCYS | 20 / 5 | 50 | 0.045 | 0.090 | 0.068 | $2.13 \cdot 10^{-11}$ |
| 500 | N4SR-SENE | 18.5 / 4 | 50 | 0.060 | 0.081 | 0.077 | $6.60 \cdot 10^{-13}$ |
| 500 | EQL÷ | 45 / 15 | 50 | 0.550 | 6.090 | 4.340 | $4.53 \cdot 10^{-17}$ |
| 500 | mSNGP-LS | NA | 50 | 0.012 | 0.036 | 0.027 | $1.26 \cdot 10^{-8}$ |
| 10 | **N4SR−ACYE** | 12 / 3 | 49 | 0.078 | 0.083 | 0.081 | − |
| 10 | N4SR-ACYS | 10 / 3 | 50 | 0.055 | 0.077 | 0.066 | $1.10 \cdot 10^{-1}$ |
| 10 | N4SR-AEYE | 12 / 3 | 47 | 0.090 | 0.093 | 0.088 | $1.0 \cdot 10^{-1}$ |
| 10 | N4SR-AEYS | 11 / 3 | 50 | 0.099 | 0.101 | 0.099 | $1.23 \cdot 10^{-1}$ |
| 10 | N4SR-AENE | 22 / 4 | 50 | 1.321 | 2.318 | 2.730 | $4.67 \cdot 10^{-13}$ |
| 10 | N4SR-SCYE | 29.5 / 5 | 50 | 0.841 | 0.942 | 1.036 | $5.75 \cdot 10^{-10}$ |
| 10 | N4SR-SCYS | 29 / 6 | 50 | 0.834 | 1.188 | 1.425 | $2.01 \cdot 10^{-10}$ |
| 10 | N4SR-SENE | 29 / 5 | 50 | 2.826 | 2.713 | 3.481 | $1.48 \cdot 10^{-14}$ |
| 10 | EQL÷ | 45 / 15 | 50 | 5.680 | 14.060 | 12.990 | $1.49 \cdot 10^{-17}$ |
| 10 | mSNGP-LS | NA | 50 | 0.008 | 0.009 | **0.008** | $2.21 \cdot 10^{-5}$ |

### 3) RESISTORS

Results obtained on the `resistors` problem are shown in Table 3 for the general architecture and in Table 4 for the informed architecture. A clear observation is that the NN architecture matters. If the topology contains only the units that are necessary to compose the desired expression, then the `N4SR` methods produce significantly better models both in terms of the test accuracy and the model's complexity.

As for the learning strategy, `N4SR-ACYE` is significantly better than the static variants only when using the informed architecture. With the general architecture, both learning strategies are comparable. `N4SR-ACYE` using the epoch-wise strategy is significantly better than `N4SR-ACYS` using the single-epoch one only when running with the general architecture on the large data set. Again, `N4SR-ACYE` clearly outperforms all `N4SR` variants that do not use prior knowledge. Interestingly, `N4SR` variants without prior knowledge perform much better than `EQL÷`, which completely fails to find good models. Note that both methods use the extrapolation data for choosing the final model. On this problem, `N4SR-ACYE` is the overall winner when running with the informed architecture on the large data set.

Figure 8 demonstrates the performance of the median and the best w.r.t. the $RMSE_{int+ext}$ models generated by the `N4SR-ACYE` method using small and large training data sets, respectively. It shows the difference between the `N4SR` model and the reference model on the whole domain. One can see that the models are doing well on the interpolation as well as on the extrapolation domain, even when trained on the small data set. The raw analytic expression represented by the best `N4SR-ACYE` model is

$$f(r_1, r_2) = 0.657 \, (-1.765 \, ((1.120 \, r_1) \, (-1.256 \, r_2)))$$
$$/ \, (1.632 \, r_1 + 1.633 \, r_2),$$

which can be rewritten as

$$f(r_1, r_2) = 1.6324 \, r_1 \, r_2 / (1.6322 \, r_1 + 1.6325 \, r_2).$$

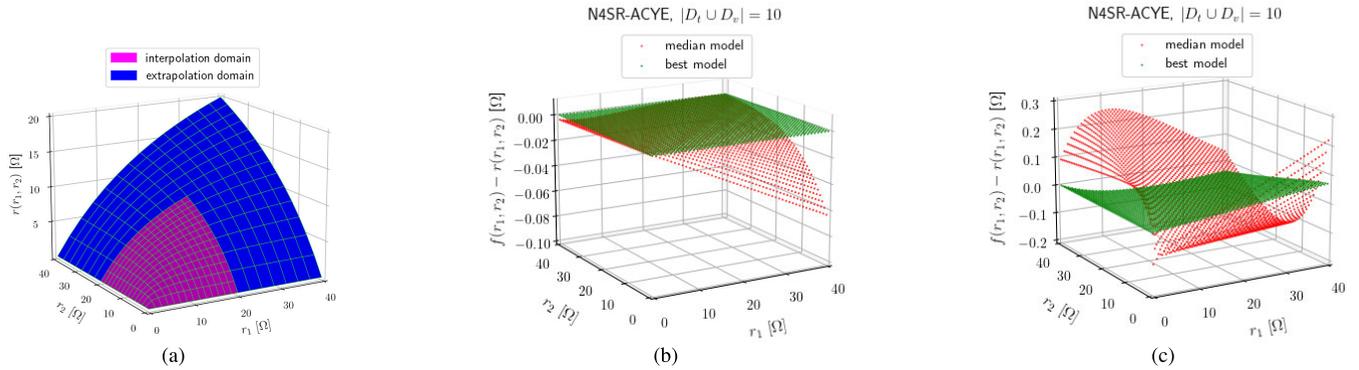

**FIGURE 8.** Examples of models generated for the `resistors` problem with `N4SR-ACYE` method. (a) is the reference model, (b) shows models attained when using the learning data set of size $|D_t \cup D_v| = 500$, and (c) shows models attained with the learning data set of size 10. Out of all best-of-run models collected over all runs for each experiment, the median model w.r.t. $RMSE_{int+ext}$ (shown in red) and the model with the best $RMSE_{int+ext}$ value (shown in green) are presented.

This is a very accurate approximation of the reference model, including the correct structure. However, this is not generally true for all output models. See more on this in Section V-F.

Figure 9 shows a typical example of one run of `N4SR-ACYE` and `N4SR-ACYS` on the `resistors` problem. It was running for 10000 iterations with the parameters' setting as described in Section V-D2. Figures 9(a) and 9(b) illustrate how the raw loss terms $\mathcal{L}^s$, $\mathcal{L}^c$, and $\mathcal{L}^r$ are continuously scaled to the values that are in the desired ratio to the values of $\mathcal{L}^t$. Note that the $\mathcal{L}^c$ and $\mathcal{L}^r$ terms are calculated starting from iteration 1000 and 2000, respectively, and $\mathcal{L}^r$ is not considered in the last 1000 iterations. Figures 9(c) and 9(d) show the effect of adaptive weighting on individual partial constraint loss terms. It can be seen that the raw values, which are of different magnitudes, are equalized throughout the whole run so that all constraints are equally important within the constraint loss term $\mathcal{L}^c$. Figures 9(e) and 9(f) show the evolution of the current model and the best-so-far one in terms of their complexity and $\mathcal{L}^t$ obtained with `N4SR-ACYE` and `N4SR-ACYS`, respectively. The initial phase up to iteration 3000 is the same for both algorithms. The rest of the run is already different for each algorithm, clearly demonstrating the effect of perturbing the current model within the exploration phase at the beginning of each epoch. One can see that the exploration introduces only a certain number of new active weights into the model, i.e., not all learnable weights are activated. `N4SR-ACYE` converges to a simpler model

(17 active weights) than `N4SR-ACYS` (22 active weights) while both models perform comparably with respect to $\mathcal{L}^t$.

Figure 10 shows the evolution of the individual constraint loss terms of the best-so-far model observed in the run with `N4SR-ACYE` from Figure 9. It can be seen that the final model perfectly satisfies the "less than input" constraint and its error with respect to the other constraints is in the order of $10^{-4}$ to $10^{-3}$.

#### 4) ANTI-LOCK BRAKING SYSTEM – MAGIC FORMULA

The results obtained on the `magic` problem are summarized in Table 5. We can see that there are no statistically significant differences between the `N4SR` variants in terms of the $RMSE_{int+ext}$ value. Even the static variants perform very well. This might indicate that setting up the loss terms weights just once for the whole run was sufficient in this case. Another observation is that the models generated by the static `N4SR` variants are roughly twice as complex as the models generated by the adaptive ones. However, this comes at the cost of a significantly lower number of non-trivial models produced by the adaptive variants. The overall best method is again `mSNGP-LS` while `EQL÷` is clearly the worst-performing method in terms of both the model performance and its complexity.

Figure 11 demonstrates the positive effect of the use of prior knowledge. We can see that both the median as well as the best model obtained with `N4SR-ACYE` and `N4SR-ACYS` comply with the constraints in the entire range of input values. For illustration, we show in detail that the models start precisely from the origin [0, 0] and all of them exhibit the positive derivative in the right part of $\mathbb{D}_i$ while monotonically decreasing to the limit value. This does not apply to either `N4SR-AENE` or `EQL÷`. These two methods do not use prior knowledge and produce models that violate the constraints. Even in this case, when the target function is simple and the training data cover the input space quite well, see Figure 5(a).

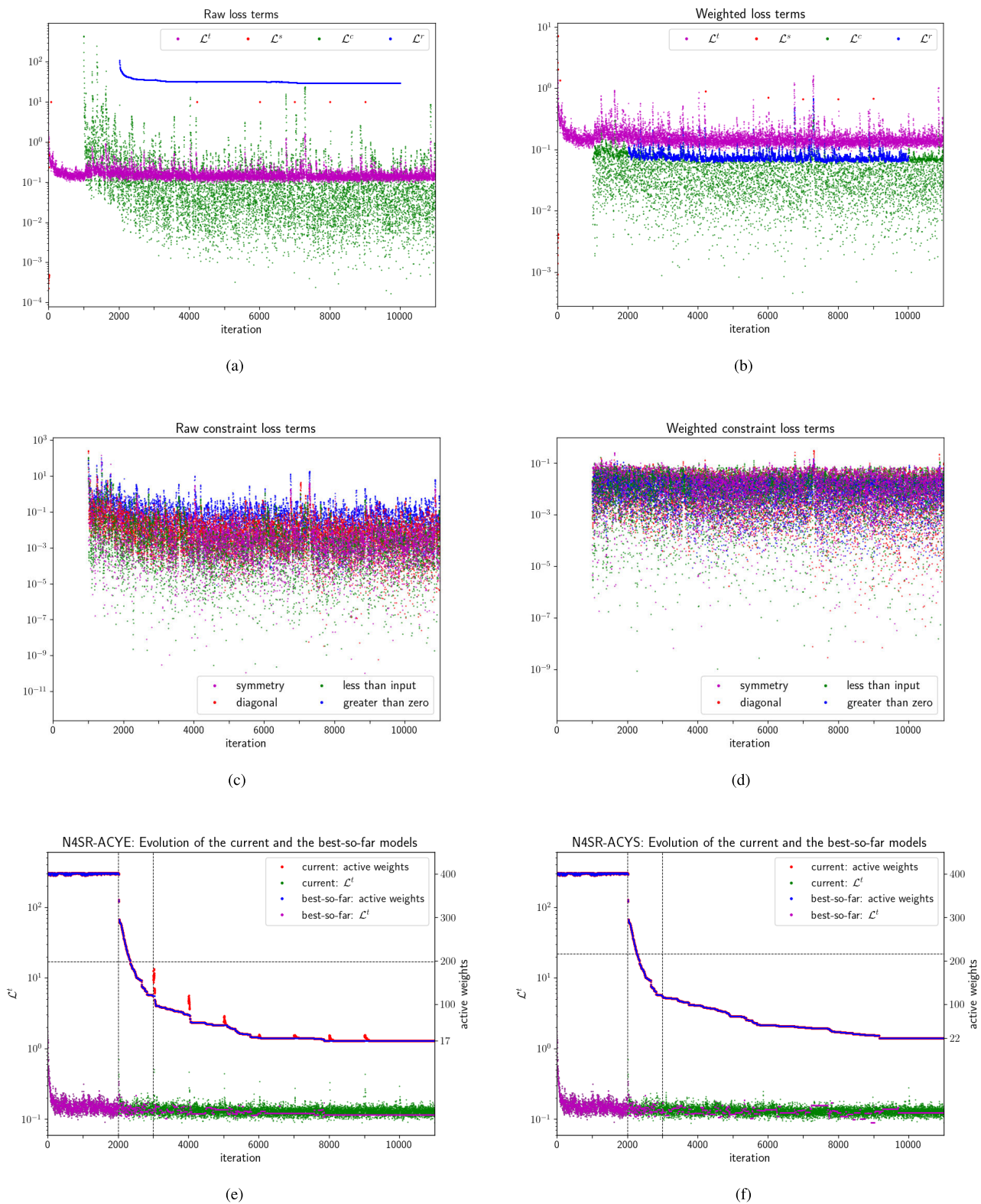

**FIGURE 9.** A typical run of `N4SR-ACYE` and `N4SR-ACYS` on the large data set of the `resistors` problem. (a)-(d) present the raw loss terms, weighted loss terms, raw constraint loss terms, and weighted constraint terms obtained with `N4SR-ACYE`. (e) shows the evolution of the current and the best-so-far models, in terms of their complexity and $\mathcal{L}^t$, observed for `N4SR-ACYE`. (f) shows the evolution of the current and the best-so-far model observed for `N4SR-ACYS`.

**TABLE 5.** Results obtained with `EQL÷`, `mSNGP-LS`, and `N4SR` methods with the general architecture on the `magic` problem. The complexity is given as the number of active links / number of active units. The complexity and the RMSE values are medians over 50 runs. The *p*-values relate to the statistical tests calculated on the $RMSE_{int+ext}$ values.

| Method | complexity | $N_{nt}$ | $RMSE_{int}$ | $RMSE_{ext}$ | $RMSE_{int+ext}$ | $p$-value |
|---|---|---|---|---|---|---|
| N4SR-ACYE | 8.5 / 4 | 14 | $6.65 \cdot 10^{-3}$ | $5.51 \cdot 10^{-3}$ | $6.22 \cdot 10^{-3}$ | – |
| N4SR-ACYS | 11 / 4 | 13 | $1.57 \cdot 10^{-2}$ | $2.01 \cdot 10^{-2}$ | $1.74 \cdot 10^{-2}$ | $8.06 \cdot 10^{-2}$ |
| N4SR-AEYE | 8.5 / 4 | 14 | $5.90 \cdot 10^{-3}$ | $3.40 \cdot 10^{-3}$ | $5.27 \cdot 10^{-3}$ | $5.20 \cdot 10^{-1}$ |
| N4SR-AEYS | 11 / 4 | 21 | $2.30 \cdot 10^{-2}$ | $2.63 \cdot 10^{-2}$ | $2.41 \cdot 10^{-2}$ | $3.81 \cdot 10^{-1}$ |
| N4SR-AENE | 18 / 5 | 50 | $7.96 \cdot 10^{-3}$ | $1.30 \cdot 10^{-2}$ | $1.06 \cdot 10^{-2}$ | $9.74 \cdot 10^{-1}$ |
| N4SR-SCYE | 19 / 5 | 50 | $4.41 \cdot 10^{-3}$ | $7.72 \cdot 10^{-3}$ | $5.92 \cdot 10^{-3}$ | $8.82 \cdot 10^{-1}$ |
| N4SR-SCYS | 16 / 5 | 50 | $4.31 \cdot 10^{-3}$ | $7.28 \cdot 10^{-3}$ | $5.00 \cdot 10^{-3}$ | $8.72 \cdot 10^{-1}$ |
| N4SR-SENE | 18.5 / 5 | 50 | $4.07 \cdot 10^{-3}$ | $6.70 \cdot 10^{-3}$ | $4.85 \cdot 10^{-3}$ | $7.21 \cdot 10^{-1}$ |
| EQL÷ | 29 / 17 | 50 | $5.27 \cdot 10^{-2}$ | $4.87 \cdot 10^{-2}$ | $5.26 \cdot 10^{-2}$ | $8.50 \cdot 10^{-6}$ |
| mSNGP-LS | NA | 50 | $7.26 \cdot 10^{-4}$ | $3.36 \cdot 10^{-3}$ | $2.01 \cdot 10^{-3}$ | $6.37 \cdot 10^{-3}$ |

For illustration, we show the raw analytic expression of the best `N4SR-ACYE` model

$$f(\kappa) = 0.596\,(-0.858\,\arctan(6.321\,(2.711\,(-4.5\,\kappa)))$$
$$- 1.268\,(2.711\,(-4.5\,\kappa)))/(-1.964\,(2.711\,(-4.5\,\kappa))$$
$$+ 3.173\,\kappa + 1.485)$$

and its equivalent obtained by symbolic simplification

$$f(\kappa) = (9.218\,\kappa - 0.511\,\arctan(-77.099\,\kappa))$$
$$/(27.126\,\kappa + 1.485).$$

### F. DISCUSSION

The results obtained with the proposed `N4SR-ACYE` method are very promising. We observed that the method is capable of finding sparse and accurate models that exhibit desired properties defined as the prior knowledge for the given problem. Nevertheless, we also observe that the GP-based `mSNGP-LS` approach often outperforms `N4SR-ACYE`. Here, we analyze the results and propose our hypothesis on why it is so.

A detailed inspection of the final models obtained with `mSNGP-LS` on the `turtlebot` problem revealed that the analytic expressions of the best-performing models are composed of simple elementary structures. By 'simple', we mean that the arithmetic operators and the sin and tanh functions, which are at the lower levels of the expression tree, operate on 'raw' variables (i.e., the variables are weighted by the coefficient of 1). An example is the best `mSNGP-LS` model

$$x_{pos,k+1} = 1.0011\,x_{pos,k} - 0.0833\,\sin(\phi_k - v_{f,k})$$
$$+ 0.0815\,\sin(\phi_k + v_{f,k})$$
$$+ 0.0018\,\sin(\phi_k + \sin(v_{a,k}))$$
$$+ 0.0029\,\tanh(v_{a,k} + 1.0) - 0.003$$

composed of simple elementary functions $\sin(\phi_k - v_{f,k})$, $\sin(\phi_k + v_{f,k})$, $\sin(\phi_k + \sin(v_{a,k}))$, and $\tanh(v_{a,k} + 1.0)$. Note that the `mSNGP-LS` works in a 'bottom-up' manner. It starts the search process with many of the elementary structures easily available and tries to combine them in the final expression optimally. Thus, it works with building blocks that are already there and 'just' searches for their best combination.

On the contrary, for NN-based SR approaches of the `N4SR` type, it is hard to converge to expressions like this.

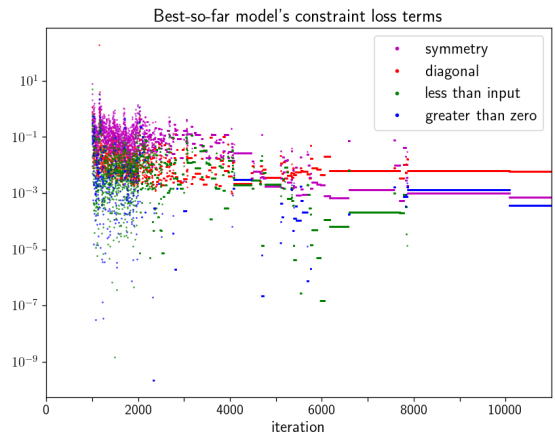

**FIGURE 10.** Evolution of individual raw constraint loss terms of the best-so-far solution during the run of `N4SR-ACYE` on the large `resistors` data set.

The method can be seen as a 'top-down' approach. It starts with the complete NN topology where each function unit has its $z$ node(s) realized as the random affine transformation of *all* previous layer's units outputs. Moreover, the learned weights are randomly initialized to rather small values, thus far from the desired value of 1. Then, it has to carefully eliminate all the useless units and useless inputs to each active function unit through many iterations of the gradient-based optimization process to get a sparse model composed of simple elementary structures. This leads to models like this:

$$x_{pos,k+1} = -0.3822\,((0.5605\,\sin(0.9838\,\phi_k + 1.5337))$$
$$(-0.7903\,v_{f,k})) + 0.999986\,x_{pos,k}.$$

This is the best `N4SR-ACYE` model on the `turtlebot` problem. We can see that the variables' multiplication coefficients are set to values very close to but not exactly one.

Another analysis revealed that on the `resistors` problem, `N4SR-ACYE` could hardly find the models that perfectly fit the expression of the reference model. In particular, only 3 out of 50 runs of `N4SR-ACYS` converged to the model that after simplification represents the expression $f(r_1, r_2) = \frac{c_1 r_1 r_2}{(c_2 r_1 + c_3 r_2)}$ where $c_1 \doteq c_2 \doteq c_3$. `N4SR-ACYE` failed to find this expression at all. Most of the time, expressions of the form $f(r_1, r_2) = \frac{c_1 r_1^2 + c_2 r_1 r_2 - c_3 r_2^2}{c_4 r_1 + c_5 r_2}$ with $c_1 \doteq c_3$, $c_2 \doteq c_4 \doteq c_5$, and $c_1 \ll c_2$ were found, which represent a close approximation of the reference model. This can also be caused by the top-down nature of the `N4SR-ACYE` approach. It may be that the sub-optimal model is more easily attainable since there are many more ways leading from the complete initial NN architecture to the sub-optimal model than to the optimal one.

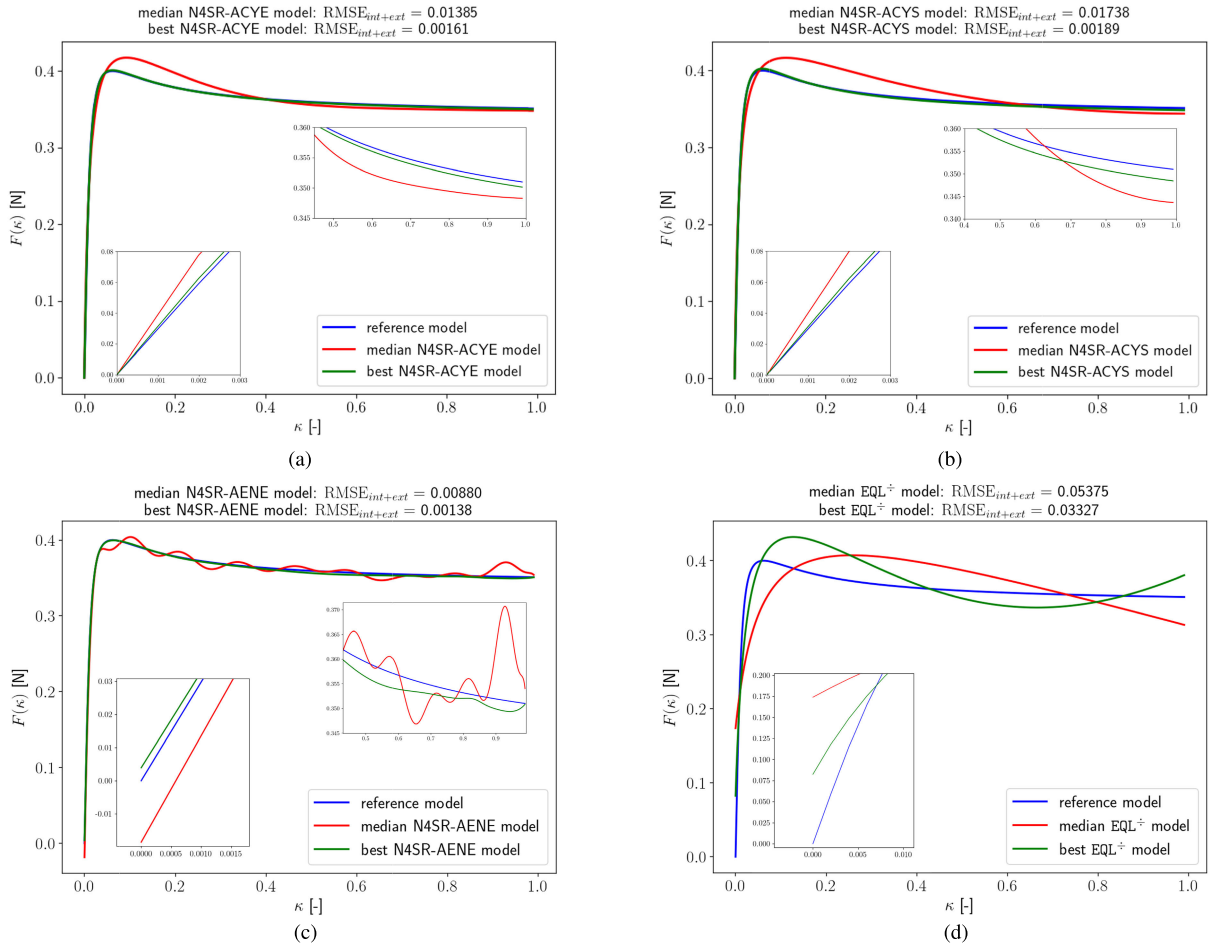

**FIGURE 11.** Examples of models generated for the `magic` problem with `N4SR-ACYE` (a), `N4SR-ACYS` (b), `N4SR-AENE` (c), and `EQL`$^{\div}$ (d) method. Out of all best-of-run models collected over all runs for each experiment, the median model w.r.t. $RMSE_{int+ext}$ (shown in red) and the model with the best $RMSE_{int+ext}$ value (shown in green) are presented.

Based on these analyses and observations, we hypothesize that when solving problems with the optimal solution composed of simple elementary structures, the `mSNGP-LS` method working in a bottom-up manner can much more efficiently arrive at the desired model than the `N4SR-ACYE` method.

## VI. CONCLUSION AND FUTURE WORK

We proposed a new neural network-based symbolic regression method, `N4SR`, that generates models in the form of sparse analytic expressions. The novelty of this approach is that it allows the incorporation of prior knowledge describing desired properties of the system to be incorporated into the learning process. It also involves components to facilitate convergence to a sparse model and to eliminate the probability of getting stuck in some poor local optimum, namely the adaptive loss terms weighting scheme and the epoch-wise learning process. Also important is the proposed method for selecting the final model of the run.

We experimentally tested the approach on four test systems and compared it to another neural network-based algorithm and to a genetic programming-based algorithm. The results clearly demonstrate the potential of the proposed method to find sparse, accurate, and physically plausible models also in cases when only a very small training data set is available. Moreover, we demonstrated that the proposed parameter-free method for the final model selection is good at identifying models that perform well in the interpolation domain (i.e., the domain from which the training data were sampled) as well as in the extrapolation domain (i.e., the domain that was completely omitted in the training data).

We also identified weaknesses of the approach, particularly its inefficiency in pruning the initial NN architecture towards the sparse one representing the final simple analytic expression. We showed that on some problems, the learning process tends to converge to suboptimal expressions due to its top-down search strategy.

In our future research, we will primarily investigate new strategies for better exploration of the space of sparse model architectures. We will focus on approaches that learn the most significant structures on the fly and use the information to guide the learning process. Other approaches will consider alternative representations of the NN units that would facilitate the pruning of the NN architecture towards a sparse final model.

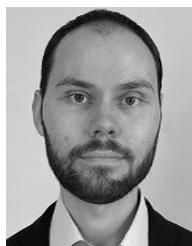

**ERIK DERNER** received the M.Sc. degree (Hons.) in artificial intelligence and computer vision and the Ph.D. degree in control engineering and robotics from Czech Technical University (CTU) in Prague, Czech Republic, in 2022. His research interests include human-centric artificial intelligence, large language models, sample-efficient model learning, genetic algorithms, and computer vision. His current research interests include safety, security, and the ethical aspects of generative and conversational artificial intelligence.

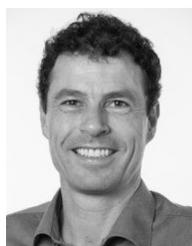

**ROBERT BABUŠKA** (Member, IEEE) received the M.Sc. degree (Hons.) in control engineering from Czech Technical University in Prague, in 1990, and the Ph.D. degree (cum laude) from the Delft University of Technology (TU Delft), The Netherlands, in 1997. He has had faculty member appointments with Czech Technical University in Prague and the Electrical Engineering Faculty, TU Delft, where he is currently a Full Professor in intelligent control and robotics with the Department of Cognitive Robotics, Faculty 3mE. In the past, he made seminal contributions to the field of nonlinear control and identification with the use of fuzzy modeling techniques. His current research interests include reinforcement learning, adaptive and learning robot control, nonlinear system identification, and state-estimation. He has been involved in the applications of these techniques in various fields, ranging from process control to robotics and aerospace.

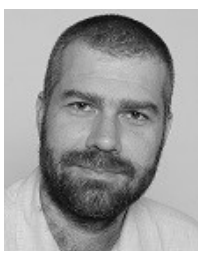

**JIŘÍ KUBALÍK** received the M.Sc. degree in computer science and the Ph.D. degree in artificial intelligence and biocybernetics from Czech Technical University (CTU) in Prague, in 1994 and 2001, respectively. He is currently a Senior Researcher with the Czech Institute of Informatics, Robotics and Cybernetics, CTU in Prague. His research interest includes evolutionary computation techniques and their applications to hard optimization problems. He is the coauthor of more than 30 papers in this area.

○ ○ ○